\documentclass[11pt]{article}

\usepackage[final]{acl}

 \usepackage{microtype}

\usepackage{authblk}
\usepackage{graphicx}
\usepackage{booktabs} 
\usepackage{amsmath} 

\usepackage[english,bidi=default]{babel} 
\babelfont{rm}{TeXGyreTermesX} 

\usepackage{multirow}
\usepackage{xcolor}

\newcommand{\drop}[1]{{\color{red!70!black}#1}}
\newcommand{\gainhigh}[1]{\textcolor{green!60!black}{#1}}

\makeatletter
\renewcommand\@makefntext[1]{\noindent\@makefnmark\,#1}
\makeatother

\title{DataKernelBench: Can LLMs Optimize Database Queries on GPUs?}

\author[1,2]{Gokul Karthik Kumar}
\author[1]{Yotam Perlitz}
\author[1]{Corey Lammie}
\author[1]{\\Andrea Giovannini}
\author[2]{Katja Hose}

\affil[1]{IBM Research, Zurich, Switzerland}
\affil[2]{TU Wien, Vienna, Austria}
\affil[ ]{\texttt{gok@zurich.ibm.com, y.perlitz@ibm.com, corey.lammie@ibm.com,}} 
\affil[ ]{\texttt{agv@zurich.ibm.com, katja.hose@tuwien.ac.at}}

\begin{document}

\maketitle

\begin{abstract}
GPUs increasingly accelerate database systems, but query-specific peak performance still often relies on hand-written kernels. Existing LLM kernel benchmarks focus on machine learning operators, leaving irregular, heterogeneous, data-movement-heavy database-style operators untested. We introduce \textbf{DataKernelBench}\footnote{\href{https://kerneldf.github.io/datakernelbench}{https://kerneldf.github.io/datakernelbench}}, which translates SQL into validated PyTorch \textbf{TorchPlan} programs and evaluates LLMs that optimize either the core tensor-bounded snippet or the full query in CUDA or Triton through execution-guided repair. Across ten proprietary and open-weight models on \textbf{TPC-H SF10} with an H100 GPU, the strongest full-query CUDA configuration achieves \textbf{2.11$\times$ speedup over \texttt{torch.compile}} at full pass rate. We find that higher-performing implementations commonly use kernel fusion and execution-strategy changes, stronger models benefit most from full-query specialization, and workload context matters more than hardware context. To handle data larger than GPU memory, we extend TorchPlan with Dask-cuDF for on-demand partition loading on \textbf{TPC-H SF100} with four H100 GPUs, achieving \textbf{2.54$\times$} speedup.
\end{abstract}

\begin{figure}[t]
    \centering
    \includegraphics[width=0.85\columnwidth]{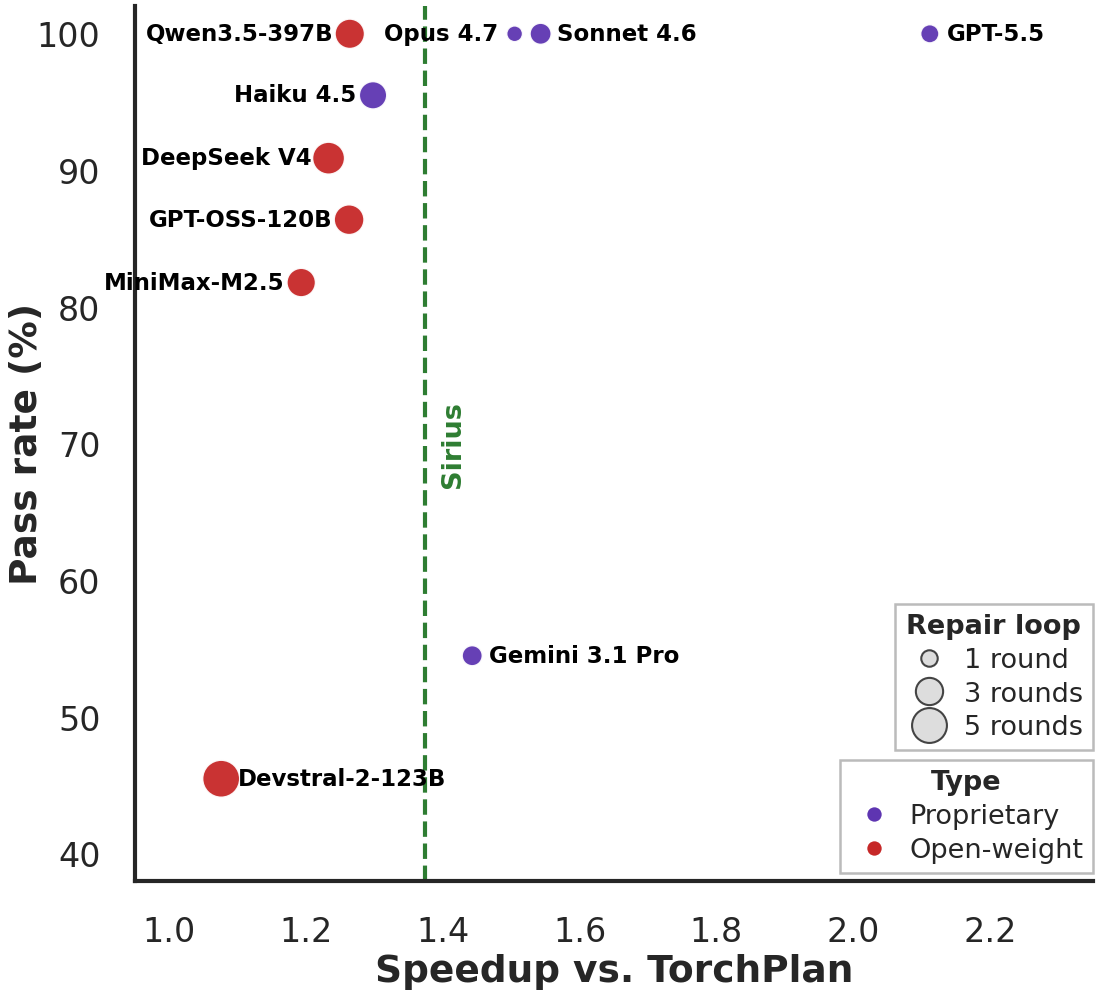}
    \caption{DataKernelBench tests LLMs on GPU kernel generation for database query optimization, where multiple models achieve 100\% pass rate and substantial speedups over the TorchPlan baseline on TPC-H SF10 with NVIDIA H100. See Table~\ref{tab:main_results} and Section~\ref{sec:overall_results}.}
\end{figure}

\section{Introduction}
Massive AI infrastructure investments \citep{idc_ai_infrastructure_2025} have made GPUs increasingly common in enterprise environments \citep{gartner_ai_spending_2025}. At the same time, many analytical workloads have execution patterns that can benefit from GPU hardware: large scans, predicate evaluation, joins, and aggregations over columnar data expose substantial data parallelism and place heavy demand on memory bandwidth. This combination makes GPU acceleration an increasingly attractive direction for analytical data processing \citep{sirius}, especially for frequently executed queries whose cost can justify query-specific optimization.

One practical approach is Tensor Query Processing (TQP) \citep{tqp1,tqp2}, which maps relational operators such as joins, filters, and aggregations to tensor computations so that database workloads can run on mature machine learning (ML) frameworks such as PyTorch. Recent work shows that this approach supports terabyte-scale analytics on modern multi-GPU systems \citep{tqp_multi_gpu}. These trends raise a natural question: if analytical queries can be expressed as tensor programs, \textit{can latest code LLMs synthesize specialized GPU kernels that outperform generic compilation?}

This question matters because generic ML compilers often fall short of peak performance on analytical workloads. Although \texttt{torch.compile} is effective for regular, compute-bound neural operators, analytical queries often involve irregular memory access, complex predicates, heterogeneous operator combinations, and query-specific fusion opportunities. Peak performance therefore still often requires hand-written bespoke GPU kernels, yet writing and maintaining them is expensive even when specialization pays off for frequently executed reporting queries \citep{lqs, bespoke_olap}.

Recent progress in LLMs \citep{openai_gpt4, anthropic_claude, qwen3_coder_next, team2024falcon, livecodebench, evalplusperformance} suggests a promising path toward reducing this engineering burden. Prior work shows that LLMs can generate specialized CUDA and Triton kernels for hardware acceleration \citep{tritonrl, cuda_kernel_agent, kernelevolve, lange2025towards, baronio2026kevin}, and recent benchmarks such as KernelBench \citep{kernelbench}, TritonBench \citep{tritonbench}, and MultiKernelBench \citep{kernelbench} evaluate this capability on ML operators. However, these benchmarks focus on regular tensor computations with predictable structure. Analytical queries can instead involve substantial data movement, combine multiple relational operators, and require strict correctness under data-dependent control flow. Strong performance on ML-centric kernel synthesis benchmarks therefore does not establish that LLMs can handle database-style workloads.

To address this gap, we introduce \textbf{DataKernelBench} for evaluating LLM-generated GPU kernels for analytical query processing. For each query, we first translate SQL into a validated PyTorch tensor program, which we call \textbf{TorchPlan}. TorchPlan serves as a benchmarkable intermediate representation: it preserves SQL semantics while exposing a stable optimization target for kernel synthesis. It separates table handling in \texttt{run\_query} from the tensor-intensive hot path in \texttt{\_query\_core}, enabling controlled specialization at two levels: \texttt{core}, where only the hot path may be replaced, and \texttt{full}, where the full internal query implementation may be rewritten while preserving the external API. Each baseline TorchPlan is validated by comparing its outputs to DuckDB \citep{duckdb} before kernel generation. We then ask LLMs to inject optimized Triton or CUDA kernels under a multi-round execution-guided repair loop.

Our primary baseline is compiled TorchPlan execution, since TorchPlan provides a validated tensor reference and \texttt{torch.compile} is the strongest generic baseline within this execution model. The benchmark therefore measures the incremental gain of bespoke LLM-generated kernels over a strong compiled tensor baseline. We also compare against external systems such as Sirius \citep{sirius} and DuckDB, but these answer a different systems question: \textit{how fast does an independent execution engine run the workload?} Sirius is best viewed as a generic GPU-specialized database system that provides drop-in GPU acceleration for existing CPU database engines, whereas our method targets bespoke kernels for specific, frequently executed queries.

DataKernelBench targets recurring queries such as scheduled reports and dashboard refreshes, where the same query template runs on refreshed data, with different user-supplied parameters, or both. Before specialization, a deployment can compare predicted generation cost with the expected reduction in execution time or compute cost over future runs. A generated kernel is adopted only after validation and when the expected savings justify specialization; otherwise, execution falls back to compiled TorchPlan or a general-purpose GPU query engine. Table~\ref{tab:main_results} reports one such runtime threshold through $N_{\mathrm{save1h}}$. As one possible integration path, a generated CUDA kernel could be wrapped as a GPU operator in an extensible engine such as Velox's experimental cuDF backend, whose \texttt{DriverAdapter} supports operator replacement, fusion, and addition \citep{velox}.

With DataKernelBench, we evaluate ten proprietary and open-weight LLMs across CUDA and Triton backends and both optimization levels (Section~\ref{sec:experiments}). Gains are highly query-dependent, and the best LLM-optimized kernels complement rather than uniformly replace a generic GPU database system. Plan inspection attributes the largest gains to kernel fusion and execution-strategy changes that avoid intermediate materialization. We also provide preliminary evidence beyond a single GPU's memory by executing TPC-H SF100 in on-demand Dask-cuDF partitions distributed across four H100 GPUs (Section~\ref{sec:scalability}).

This work makes three contributions. 
First, we introduce, to the best of our knowledge, the first systematic benchmark for LLM-generated GPU kernels for database queries, as opposed to ML operators. 
Second, we provide a validated SQL-to-TorchPlan pipeline and modular Python TorchPlan programs that expose an LLM-friendly optimization target and serve as pluggable reference implementations for kernel synthesis.
Third, our evaluation of ten LLMs characterizes how model strength, optimization scope, prompt context, programming interface, and query structure affect correctness and performance, and uses plan-level comparisons to explain where the largest gains arise.

\section{Related Work}

\subsection{Databases, GPUs, and Analytical Processing}

GPU-accelerated analytical processing has been pursued through modular libraries and full database systems. Libraries such as \texttt{libcudf} and RAPIDS cuDF \citep{cudf} accelerate dataframe-style workloads \citep{pandas}, while composable engines such as Velox \citep{velox} explore how specialized operator libraries can be integrated into flexible pipelines. Other work develops GPU-specialized database systems such as Sirius \citep{sirius}, Kinetica \citep{kinetica}, and SQream \citep{sqream}. At a lower level, Kernel Weaver \citep{kernel_weaver} shows that automatically fused GPU kernels can substantially reduce redundant data movement in relational workloads. These results reinforce the value of specialization for analytical GPU processing, but they operate at the level of a full execution system, compiler framework, or fixed relational primitives. In contrast, our focus is to evaluate whether LLMs can synthesize \emph{bespoke kernels} for analytical queries within an existing tensor-based query pipeline.

Our work is closest in spirit to Tensor Query Processing (TQP) \citep{tqp1,tqp2}, which maps relational operators to tensor computations on ML runtimes. Subsequent work narrows the mismatch between SQL operators and tensor operations \citep{tqex} and supports execution directly on compressed data \citep{tqp_compressed}. Recent work also shows that tensor-based query processing can scale to terabyte-scale multi-GPU analytics \citep{tqp_multi_gpu} and distributed, storage-resident OLAP \citep{pystachio}. However, compiled tensor execution does not eliminate the need for specialization: analytical queries often involve irregular access patterns, heterogeneous operators, and query-specific fusion opportunities that expose a generalization ceiling in generic compilation. DataKernelBench is designed to evaluate whether LLM-generated kernels can close this gap.

\subsection{AI and GPUs for Kernel Synthesis}

Recent progress in code LLMs has led to advances in automated kernel generation for hardware acceleration \citep{openai_gpt4, anthropic_claude, qwen3_coder_next, glm}. Systems such as TritonRL \citep{tritonrl}, CUDA Agent \citep{cuda_kernel_agent}, and KernelEvolve \citep{kernelevolve} iteratively synthesize and optimize CUDA or Triton kernels, showing that LLM-based kernel generation can be a viable path toward hardware specialization. Several benchmarks now evaluate this capability systematically. KernelBench \citep{kernelbench}, TritonBench \citep{tritonbench}, and MultiKernelBench \citep{multikernelbench} measure pass rate and performance for LLM-generated kernels across hardware targets and programming abstractions.

These benchmarks are an important methodological foundation for our work, but they are centered on machine learning operators, where computations are typically dense, regular, and dominated by homogeneous tensor primitives. Analytical query processing instead combines heterogeneous relational operators with irregular data access and strict output semantics. DataKernelBench builds on the evaluation style of these prior benchmarks while shifting the target domain from ML kernels to database-style analytical workloads.

\subsection{AI and Data Systems}

AI has long been applied to data systems, first by replacing hand-designed heuristics with learned components such as learned query optimizers \citep{neo_optimizer} and learned indexes \citep{learned_index}.
With the rise of LLMs, AI in data systems has expanded from user-facing interfaces such as text-to-SQL \citep{bird_text2sql, dail_sql} toward automated system synthesis. The closest line of work to ours is Bespoke OLAP \citep{bespoke_olap}, which synthesizes workload-specific database engines to outperform general-purpose systems such as DuckDB \citep{duckdb}. Our work differs in both scope and interface: rather than generating a standalone database engine, we study whether LLMs can synthesize \emph{pluggable GPU kernels} that accelerate specific bottlenecks within validated TorchPlan pipelines.

\begin{figure*}[t]
    \centering
    \includegraphics[width=1\textwidth]{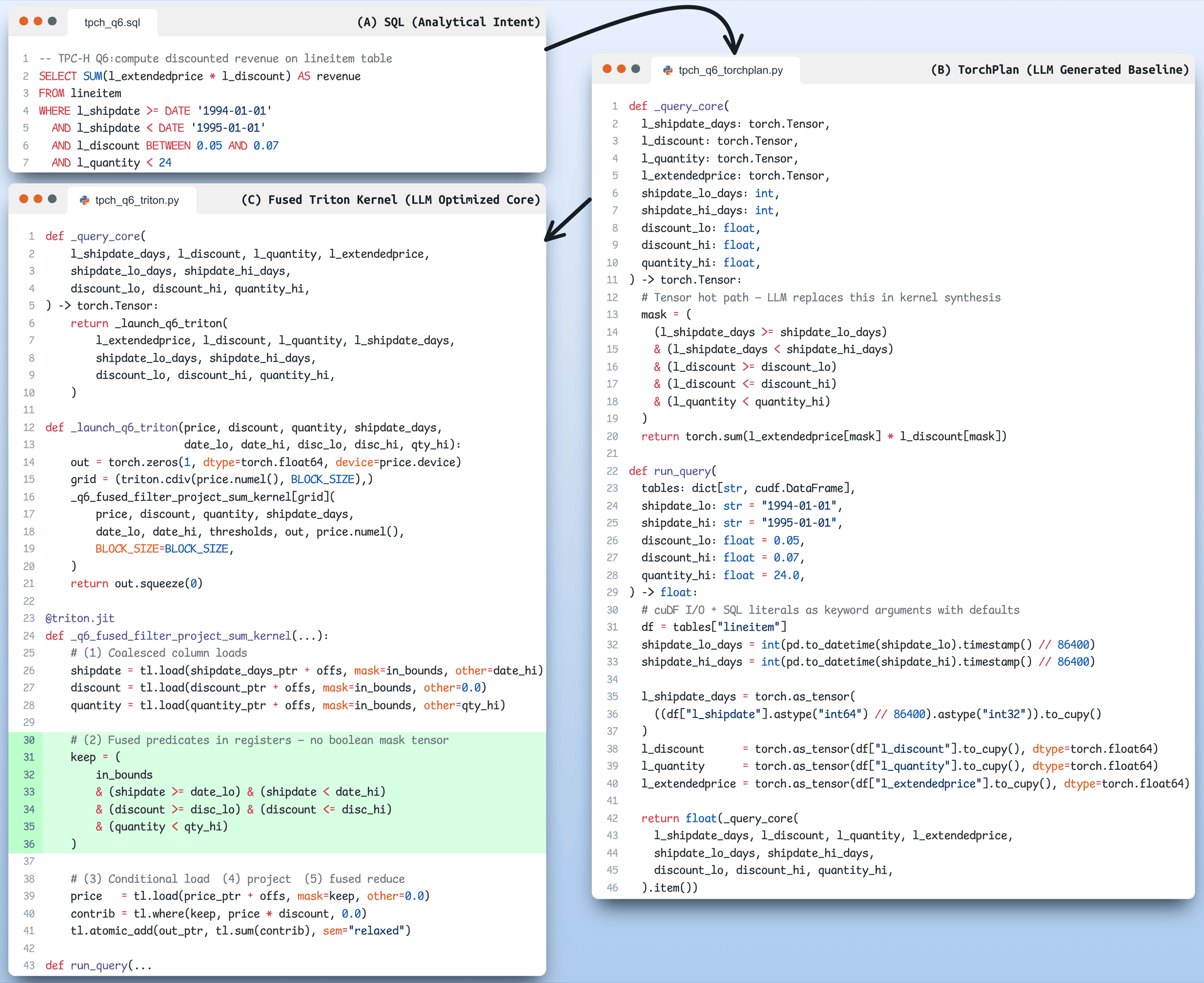}
    \caption{From SQL to TorchPlan to an LLM-synthesized fused GPU kernel for TPC-H Q6. \textbf{(A)} SQL with parameterized literals. \textbf{(B)} TorchPlan, where \texttt{run\_query} handles cuDF table processing and \texttt{\_query\_core} defines the tensor hot path optimized by \texttt{torch.compile} in the baseline. \textbf{(C)} An LLM-generated fused Triton implementation at the \texttt{core} level. The analogous CUDA example appears in Figure~\ref{app:fig:translation_cuda} of Appendix~\ref{app:cuda_example}; the Dask-cuDF variant of (B), as described in Section~\ref{sec:scalability}, appears in Figure~\ref{app:fig:translation_dask} of Appendix~\ref{app:dask_example}.}
    \label{fig:translation}
\end{figure*}

\section{The DataKernelBench Framework}
\label{sec:framework}

DataKernelBench instantiates LLM-based kernel synthesis for analytical query processing as a two-stage benchmark pipeline: (1) construct and validate a baseline \textbf{TorchPlan} for each query, and (2) ask LLMs to inject optimized Triton or CUDA kernels that preserve semantics while improving runtime. Figure~\ref{fig:translation} illustrates this SQL$\rightarrow$TorchPlan$\rightarrow$kernel mapping for TPC-H Q6. While Q6 is a simple single-table query that we chose to illustrate our approach, our benchmark covers all 22 TPC-H queries, including joins, nested subqueries, and more complex hot paths. This design cleanly separates benchmark construction from model evaluation: TorchPlans define validated task instances, while kernel synthesis measures how effectively LLMs optimize those instances under fixed pass rate and runtime constraints. The supplementary material includes the prompts, validated TorchPlans, and selected optimized CUDA and Triton implementations. We will release these artifacts and the evaluation harness at \url{https://github.com/kerneldf/datakernelbench}.

\subsection{Benchmark Workload and Setting}
\label{sec:workload}

Our primary evaluation uses \textbf{TPC-H}~\citep{tpch} at \textbf{scale factor 10} which contains 22 analytical queries over 8 relational tables. We generate benchmark artifacts using DuckDB's \texttt{tpch} extension, including data, SQL queries, and reference outputs. At runtime, tables are loaded into \texttt{cuDF}~\citep{cudf} dataframes so that \texttt{run\_query} executes on GPU-resident data.

\subsection{TorchPlan: A Benchmarkable Intermediate Representation}
\label{sec:torchplan}

Each query is first translated by an LLM into a \textbf{TorchPlan}, an executable PyTorch tensor program that serves as the benchmark's intermediate representation (IR) within the TQP paradigm. TorchPlan preserves SQL semantics while exposing a stable optimization target for kernel synthesis. In this sense, TorchPlan acts as a \emph{verified contract} between declarative query logic and imperative GPU kernel generation: it is directly executable, amenable to differential testing against a reference engine, and compatible with multiple optimization backends. We do not propose a new TQP system; rather, we use validated TQP-style tensor programs as a strong reference representation and execution substrate, then measure how much additional performance bespoke LLM-generated kernels can obtain beyond compiled tensor execution. Each TorchPlan contains two functions:
\begin{itemize}
    \item \texttt{run\_query(tables: dict[str, cudf.DataFrame], ...)}, which handles table access, joins, projections, parameter parsing, and output formatting; and
    \item \texttt{\_query\_core(...)}, which implements the tensor-intensive hot path over aligned 1D GPU tensors and Python scalars.
\end{itemize}
This decomposition mirrors the TQP design: relational preparation remains in the dataframe layer, while the tensor-intensive hot path is isolated in \texttt{\_query\_core} for specialization. TorchPlan also reflects realistic analytical execution, where the same query template may be run repeatedly with different literals. Accordingly, \texttt{run\_query} exposes SQL literals as keyword arguments with defaults, while \texttt{\_query\_core} receives only aligned GPU tensors and numeric scalars. Figure~\ref{app:fig:multi_join_torchplan} in Appendix~\ref{app:multi_join_example} provides a representative multi-join example that makes this boundary concrete.

We generate baseline TorchPlans with a single LLM (Claude Opus 4.7 in our main setup) using a fixed prompt template provided in the supplementary material. We then validate each generated plan by executing the original SQL in DuckDB and comparing the result to \texttt{run\_query} on the same input tables. Only validated TorchPlans enter the benchmark, and each is held fixed across all model, framework, and optimization-level evaluations. These TorchPlans serve as both the correctness reference for synthesized kernels and the baseline execution path from which speedup is measured. To assess how the TorchPlan generator affects final optimization, Appendix~\ref{app:torchplan_sensitivity} repeats the evaluation with GPT-5.5-generated TorchPlans. The key findings persist, although absolute runtimes change.

\subsection{Kernel Synthesis Task}
\label{sec:kernel_task}

Given a validated TorchPlan, the benchmark asks an evaluated LLM to produce a faster implementation that preserves the external \texttt{run\_query} semantics. We study two optimization levels:
\begin{itemize}
    \item \texttt{core}: the model may modify only \texttt{\_query\_core} and private helper functions; \texttt{run\_query} must remain unchanged. This keeps the optimization target close to ML kernel benchmarks, whose inputs and outputs are tensors and scalars rather than complex datatypes (e.g., String and DataFrames);
    \item \texttt{full}: the model may rewrite \texttt{run\_query}, \texttt{\_query\_core}, and helpers, while preserving the external \texttt{run\_query} API.
\end{itemize}
We support two GPU programming interfaces, \textbf{Triton} and \textbf{CUDA}. In both cases, the model must return one complete Python module exposing \texttt{run\_query}. This turns kernel synthesis into a constrained program-rewriting task: the model is free to optimize the internal computation, but not to alter the benchmark-facing semantics.

\subsection{Prompt Design}
\label{sec:prompt}

Kernel-synthesis prompts are composed from six markdown blocks concatenated in fixed order: \textsc{Base}, \textsc{Level}, \textsc{Framework}, \textsc{Gpu}, \textsc{Env}, and \textsc{Data}. The \textsc{Base} block defines task constraints, interface rules, and inserts the reference TorchPlan and original SQL. \textsc{Level} specifies the allowed optimization scope. \textsc{Framework} provides Triton- or CUDA-specific guidance and a worked fused-kernel example. \textsc{Gpu} and \textsc{Env} describe the target hardware and runtime environment. \textsc{Data} provides workload metadata such as table names, dtypes, row counts, and approximate memory footprint at scale factor 10. This modular design supports controlled ablations, since workload, hardware, and framework context can be removed independently.

\subsection{Correctness and Speedup Criteria}
\label{sec:correctness}

For each generated module, we compare the output of \texttt{run\_query} against the baseline TorchPlan on the fixed scale factor 10 tables. We report mismatches hierarchically: shape, column names and order, row count, and then cell-level differences with sampled offending values. Non-floating-point columns (e.g., integers, strings, and dates) must match exactly. Following prior kernel-benchmark practice~\citep{kernelbench}, floating-point columns are compared using \texttt{numpy.isclose} with \texttt{atol}=\texttt{rtol}=$10^{-4}$ for \texttt{float32} and $10^{-6}$ for \texttt{float64}. Failed checks are returned to the model as repair feedback.

We measure end-to-end \texttt{run\_query} runtime. Before timed measurement, we discard $N_{\text{warmup}}$ warmup runs to amortize compilation and GPU warm-start effects. Let $T_{\text{base}}$ denote the median runtime of the validated TorchPlan when \texttt{\_query\_core} is wrapped with \texttt{torch.compile}, and let $T_{\text{cand}}$ denote the median over $N_{\text{timed}}$ timed runs of the generated module after warmup. We define $\text{speedup}$ as $T_{\text{base}} / T_{\text{cand}}$.
A generated module is accepted only if it is functionally correct and achieves $\text{speedup} \ge s_{\text{min}}$. We require both correctness and a minimum speedup so that accepted modules strictly improve on compiled TorchPlan; otherwise the baseline remains preferable.

\subsection{Execution-Guided Generation}
\label{sec:gen_workflow}

We evaluate every combination of TPC-H query, model, framework (Triton or CUDA), and optimization level (\texttt{core} or \texttt{full}) as an independent run. For each combination, we execute a multi-round repair loop with at most $R_{\text{max}}$ rounds. In round~1, the model receives the composed prompt from Section~\ref{sec:prompt} and returns one candidate Python module exposing \texttt{run\_query}. Each later round appends feedback from the previous candidate, including validation failures, output mismatches, tracebacks, and measured speedup, and asks the model to produce a revised candidate. We stop at the first candidate that passes the correctness checks in Section~\ref{sec:correctness} and satisfies $\text{speedup} \ge s_{\text{min}}$, or when $R_{\text{max}}$ is reached.
To bound hung executions, correctness checks (Section~\ref{sec:correctness}) and warmup runs are terminated after $\max(T_{\text{min}}, M_{\text{warmup}}\cdot T_{\text{base}})$, while each timed run is terminated after $M_{\text{timed}}\cdot T_{\text{base}}$.
Defaults are: $R_{\text{max}}{=}10$, $s_{\text{min}}{=}1.05$, $N_{\text{warmup}}{=}2$, $N_{\text{timed}}{=}5$, $M_{\text{warmup}}{=}20$, $M_{\text{timed}}{=}3$, $T_{\text{min}}{=}1$~minute, and $T_{\text{wall}}{=}15$~minutes.

\begin{table*}[t]
\centering
\footnotesize
\setlength{\tabcolsep}{2pt}
\begin{tabular}{@{}lllrrrrrrrrrr@{}}
\toprule
\textbf{Model} & \textbf{Best} &
\textbf{Pass} &
\multicolumn{4}{c}{\textbf{$\uparrow~$Speedup vs.\ TorchPlan}} &
\multicolumn{2}{c}{\textbf{$\downarrow~$Runtime}} &
\multicolumn{4}{c}{\textbf{$\downarrow~$Mean Generation Cost}} \\
\cmidrule(lr){4-7}\cmidrule(lr){8-9}\cmidrule(lr){10-13}
 &\textbf{Config} &\textbf{Rate\%} &
 \textbf{Overall} &
\textbf{$\geq$1.05$\times$} & \textbf{$\geq$1.20$\times$} & \textbf{$\geq$2$\times$} &
\textbf{Overall (s)} & $\mathbf{N_{\mathrm{save1h}}}$ &
$\mathbf{R}$ & \textbf{In (k)} & \textbf{Out (k)} & \textbf{USD} \\
\midrule
\multicolumn{13}{@{}l}{\textit{Proprietary models}} \\
\midrule
GPT-5.5           & CUDA-full   & \textbf{100.0} & \textbf{2.11} & \textbf{100.0} &  \textbf{95.5} & \textbf{45.5} & \textbf{0.71} & \textbf{4535} & \underline{1.4} & \textbf{12.6} &  9.8 & 0.18 \\
Claude Sonnet 4.6 & Triton-full & \textbf{100.0} & \underline{1.54} &  \underline{95.5} &  77.3 & 13.6 & \underline{0.98} & \underline{6786} & 1.9 & 26.0 &  \underline{6.2} & \underline{0.13} \\
Claude Opus 4.7   & CUDA-full   & \textbf{100.0} & 1.51 & \textbf{100.0} &  \underline{81.8} & 13.6 & 1.00 & 7116 & \textbf{1.0} & 14.0 &  \textbf{4.7} & 0.14 \\
Gemini 3.1 Pro    & CUDA-full   &  54.5 & 1.44 &  50.0 &  45.5 & \underline{18.2} & 1.04 & 7775 & 1.7 & \underline{13.0} & 14.9 & 0.21 \\
Claude Haiku 4.5  & Triton-full &  \underline{95.5} & 1.30 &  90.9 &  68.2 &  9.1 & 1.16 & 10394 & 3.2 & 59.9 &  9.1 & \textbf{0.08} \\
\midrule
\multicolumn{13}{@{}l}{\textit{Open-weight models}} \\
\midrule
Qwen3.5-397B-A17B & Triton-full & \textbf{100.0} & 1.26 &  86.4 &  72.7 &  4.5 & 1.19 & 11432 & 3.7 & 58.8 & 16.9 & --- \\
GPT-OSS-120B      & CUDA-full   &  86.4 & 1.26 &  81.8 &  72.7 &  4.5 & 1.19 & 11484 & 3.8 & 69.7 & 14.1 & --- \\
DeepSeek-V4-Flash & Triton-core &  90.9 & 1.23 &  86.4 &  63.6 &  0.0 & 1.22 & 12633 & 4.4 & 80.9 &  9.7 & --- \\
MiniMax-M2.5      & Triton-full &  81.8 & 1.19 &  81.8 &  68.2 &  4.5 & 1.26 & 14786 & 3.5 & 50.4 & 15.9 & --- \\
Devstral-2-123B   & Triton-full &  45.5 & 1.08 &  36.4 &  27.3 &  0.0 & 1.40 & 33782 & 5.9 & 124.2 & 12.0 & --- \\
\midrule
\multicolumn{13}{@{}l}{\textit{Primary baselines}} \\
\midrule
TorchPlan (compile) & --- & 100.0 & 1.00 &   0.0 &   0.0 &  0.0 & 1.51 & --- & --- & --- & --- & --- \\
TorchPlan (eager)   & --- & 100.0 & 0.88 &  13.6 &   0.0 &  0.0 & 1.71 & --- & --- & --- & --- & --- \\
\midrule
\multicolumn{13}{@{}l}{\textit{DB baselines}} \\
\midrule
Sirius (GPU)  & --- & 100.0 & 1.37 &  68.2 &  63.6 & 54.5 & 1.10 & 8765 & --- & --- & --- & --- \\
DuckDB (CPU)  & --- & 100.0 & 0.54 &   9.1 &   9.1 &  4.5 & 2.81 & --- & --- & --- & --- & --- \\
\bottomrule
\end{tabular}
\caption{DataKernelBench leaderboard results on TPC-H SF10 using an NVIDIA H100, reporting the best kernel configuration per model. \textbf{Pass Rate\%} denotes the percentage of queries matching the reference output. \textbf{Speedup} measures performance relative to TorchPlan with \texttt{torch.compile} (TC), where the threshold columns indicate the fraction of the 22 queries achieving at least $1.05\times$, $1.20\times$, and $2\times$ speedups. $\mathbf{N_{\mathrm{save1h}}}$ represents the number of repetitions required to save 1 hour of execution time relative to TC. $\mathbf{R}$  is the number of rounds used to get final generation. \textbf{In (k)} \& \textbf{Out (k)} denotes token usage in thousands; \textbf{USD} is the API usage cost. Arrows indicate higher ($\uparrow$) or lower ($\downarrow$) values are better. The best results are \textbf{bolded} and second-best are \underline{underlined}. }
\label{tab:main_results}
\end{table*}

\begin{figure*}[t]
    \centering
    \includegraphics[width=0.9\textwidth]{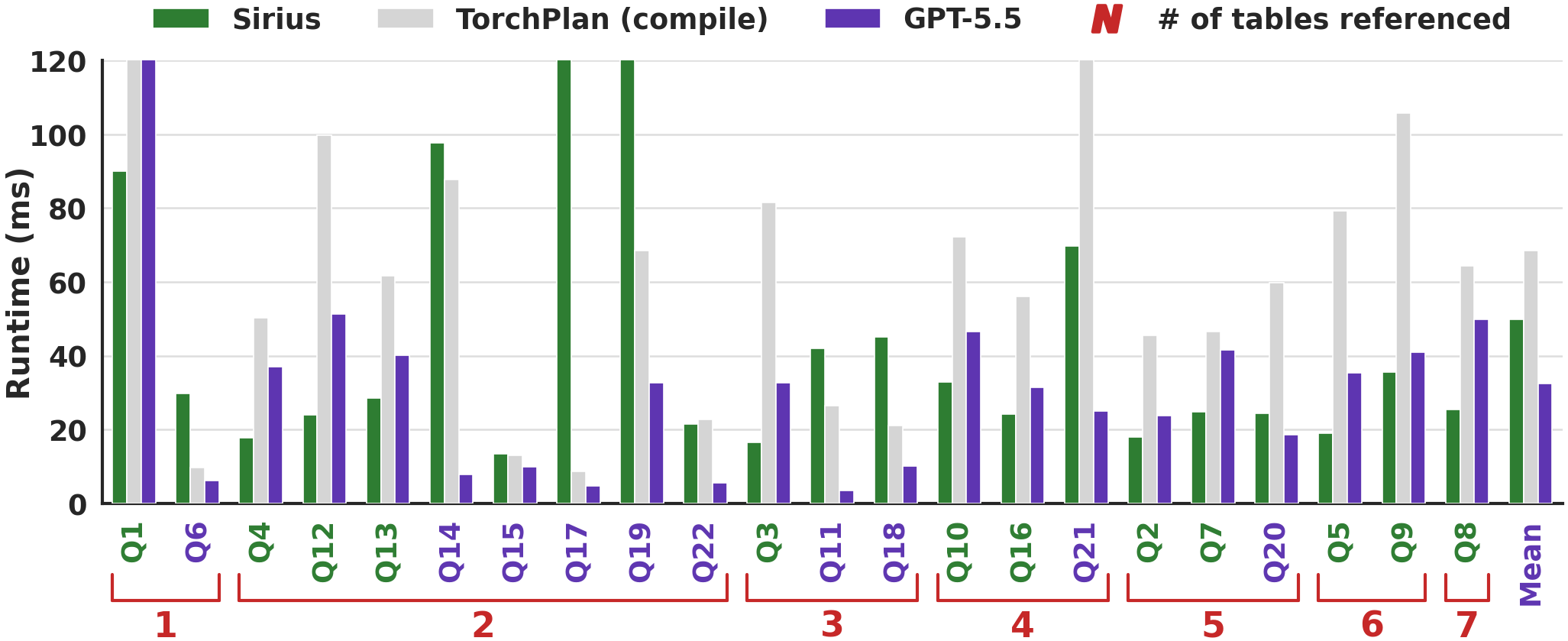}
    \caption{Per-query runtime comparison: GPT-5.5 CUDA at Full level vs. baselines. Y-axis capped at 120ms.}
    \label{fig:runtime}
\end{figure*}

\section{Experiments}
\label{sec:experiments}

\subsection{Experimental Setup}
\label{sec:exp_setup}

We evaluate ten LLMs spanning proprietary APIs and open-weight models: GPT-5.5 \citep{gpt55}, Claude Opus 4.7 \cite{anthropic2026opus47}, Claude Sonnet 4.6 \citep{anthropic2026sonnet46}, and Claude Haiku 4.5 \cite{anthropic2025haiku45}, Gemini 3.1 Pro \citep{google2026gemini31pro}, Qwen3.5-397B-A17B \citep{qwen3.5}, GPT-OSS-120B \citep{gptoss}, DeepSeek-V4-Flash \citep{deepseekv4}, MiniMax-M2.5 \citep{MiniMax-M2.5}, and Devstral-2-123B \citep{devstral}. All open-weight models were served on a separate machine. For each model, we evaluate CUDA and Triton generation, on an H100 80GB GPU, at both \texttt{core} and \texttt{full} levels, yielding 40 model--framework--level combinations. We use the default generation settings for each API or model checkpoint (e.g., temperature and top-$p$).

Within TorchPlan, \texttt{torch.compile} (TC) is the primary generic baseline, and all LLM speedups are measured against it; we also report TorchPlan eager execution to contextualize the strength of TC. For LLM-generated kernels, speedup and runtime use fallback to TC: if a generated kernel is functionally incorrect or fails to meet the minimum speedup threshold, that query is credited at the TC runtime, since such a kernel would not be adopted in practice. We also report Sirius and DuckDB as external DB baselines. Because these are independent execution engines rather than in-pipeline replacements, we always report their observed runtimes directly, including cases below parity with TC. We report the hardware and software environment details for benchmark execution, in Appendix~\ref{app:hardware-software-environment}.


\subsection{Comparative Evaluation of LLMs}
\label{sec:overall_results}

Table~\ref{tab:main_results} presents the main leaderboard on TPC-H SF10 on H100. The strongest result is GPT-5.5 with CUDA at the \texttt{full} level, which achieves \textbf{100\%} pass rate and \textbf{2.11$\times$} overall speedup over TorchPlan-\texttt{compile}. This also outperforms Sirius at \textbf{1.37$\times$}. Among open-weight models, Qwen3.5-397B-A17B with Triton-\texttt{full} is strongest at \textbf{1.26$\times$} with \textbf{100\%} pass rate, while GPT-OSS-120B matches the same speedup with lower pass rate.

The leaderboard shows a clear relationship between model quality and synthesis efficiency. Top-ranked models generally require fewer repair rounds and fewer tokens to reach their best configuration, whereas weaker models spend more on unsuccessful repairs. For example, GPT-5.5 reaches the top result with only \textbf{1.4} mean rounds, while lower-ranked open models require \textbf{3.5--5.9} rounds. At the same time, token cost alone is not a sufficient proxy for quality: Gemini 3.1 Pro reaches \textbf{1.44$\times$} speedup with moderate token cost, but passes only \textbf{54.5\%} of queries. Appendix~\ref{app:repair_behavior} analyzes all 880 generation trajectories (22 queries $\times$ 10 models $\times$ 2 frameworks $\times$ 2 optimization levels), including repair progression, failure modes, and query difficulty.

Backend preference is model-dependent. Triton is the best configuration for six of the ten models, while CUDA is best for four. Interestingly, both OpenAI models achieve their best results with CUDA, despite OpenAI's central role in Triton's development. A plausible interpretation is that Triton is easier to synthesize reliably, which benefits weaker models, while stronger models are able to exploit the higher performance ceiling of CUDA.

\subsection{Additional Correctness Validation}
\label{sec:additional_correctness}

The primary pass rate uses output matching on SF10 data, as defined in Section~\ref{sec:correctness}. We further subjected all 22 GPT-5.5 CUDA-\texttt{full} implementations, the top-performing configuration, to two complementary robustness checks beyond this protocol: source-code audits and held-out differential testing.

For the source audits, GPT-5.5 produced two verdicts for each implementation. \texttt{semantic\_match} checks predicates, join and group-by keys, aggregation formulas, date and null handling, output schema, and parameter use. \texttt{no\_cheating} checks for hard-coded outputs, cached or reference answers, and bypassed input tables. For held-out testing, we used SF1 and a perturbed SF10 dataset that preserves keys and text identifiers while applying up to $\pm30\%$ jitter to numeric columns, shifting dates by up to $\pm30$ days, and resampling low-cardinality values from their existing domains. \textbf{All 22 implementations passed both source-audit verdicts and both held-out validations}. These checks provide evidence beyond output matching on the original SF10 data, although they do not constitute a formal proof of semantic equivalence.

\subsection{Understanding Performance Gains}
\label{sec:query_level_results}
\label{sec:plan_analysis}

Figure~\ref{fig:runtime} shows per-query runtime for TorchPlan-\texttt{compile}, Sirius, and GPT-5.5 CUDA-\texttt{full}. The main pattern is strong non-uniformity: Sirius is fastest on many queries, while GPT-5.5-generated kernels dominate on a different subset. The advantage narrows as queries reference more tables. GPT-5.5 is faster than Sirius on \textbf{8/13} queries referencing at most three tables, but only \textbf{2/9} queries referencing more than three, while overall runtime still favors GPT-5.5. This pattern is consistent with Appendix~\ref{app:chokepoint-analysis}: all ten LLMs show negative correlation between choke-point coverage and speedup, whereas Sirius shows positive correlation.

To identify where the speedups arise, we compared GPT-5.5 CUDA-\texttt{full} implementations with their compiled TorchPlan baselines across all 22 queries. The dominant pattern is \textbf{kernel fusion}: filtering, projection, and aggregation are combined into one or a few passes, avoiding intermediate masks, gathers, and temporary tensors. \textbf{Hybrid execution} is also common: 21/22 implementations retain cuDF for string or DataFrame operations and use CUDA for the numeric hot path. In addition, 14/22 replace \texttt{torch.unique}/\texttt{scatter\_add} group-by pipelines with \textbf{fused CUDA aggregation}.

Q14 provides the clearest example of a broader plan rewrite. GPT-5.5 builds a compact lookup of promotional \texttt{part} keys and probes it while scanning \texttt{lineitem}, avoiding a materialized join. This yields \textbf{11.2$\times$ speedup}, while CUDA-\texttt{full} alternatives retaining the cuDF merge remain near 1.07$\times$. Similar advantages appear on Q17 and Q19. These results support a complementary interpretation: bespoke LLM-synthesized kernels do not replace generic GPU database systems on every query, but can deliver large wins on selected analytical pipelines that change the benchmark-wide ranking.


\subsection{Beyond GPU Memory: Partitioned Multi-GPU Execution}
\label{sec:scalability}

We conducted a proof of concept for data that does not fit in GPU memory by replacing eager cuDF tables, which are loaded in full, with \textbf{partitioned Dask-cuDF} tables loaded on demand \citep{dask,cudf}. The extended \texttt{run\_query} uses \texttt{map\_partitions} to invoke \texttt{\_process\_partition} on each cuDF partition, which converts its columns to tensors and calls \texttt{\_query\_core}. Figure~\ref{app:fig:translation_dask} in Appendix~\ref{app:dask_example} shows this interface for TPC-H Q6. Dask distributes these partitions across the available GPU workers, enabling the same partitioned execution to use multiple GPUs.

We generated partitioned TorchPlans and GPT-5.5 CUDA-\texttt{full} implementations for \textbf{TPC-H SF100} and evaluated them on \textbf{four H100 80\,GB GPUs} using 10\,GB chunks. \textbf{All 22} optimized implementations passed the correctness checks. The complete workload finished in 36.44\,s, a \textbf{2.54$\times$ speedup} over the corresponding Dask-cuDF TorchPlan baseline. Unlike the main SF10 evaluation, this timing includes on-demand data loading and materialization. These results provide preliminary evidence that the approach can extend to partitioned data larger than GPU memory; Appendix~\ref{app:gpu_scaling} reports how the same plans scale from 1 to 4 GPUs.

\subsection{Effect of Optimization Scope: \texttt{core} vs.\ \texttt{full}}
\label{sec:core_full_results}

Table~\ref{tab:core_full_delta} compares \texttt{full} against \texttt{core} optimization by leaderboard tier. For the top-5 models, expanding synthesis from the tensor hot path to the full TorchPlan yields mean speedup gains of \textbf{+0.31} for CUDA and \textbf{+0.29} for Triton. For the bottom-5 models, the gains are only \textbf{+0.02} and \textbf{+0.05}.

This suggests that end-to-end query specialization is an emergent capability of stronger code LLMs rather than a uniform benefit across the leaderboard. Weaker models can sometimes optimize local tensor kernels, but they obtain little additional benefit from the broader optimization surface exposed by \texttt{full} generation. In contrast, stronger models are able to exploit opportunities that span table handling, launch logic, and fused execution, which helps explain why the leaderboard is led by \texttt{full}-level configurations.

\begin{table}[t]
\centering
\small
\setlength{\tabcolsep}{6pt}
\begin{tabular}{@{}lrr@{}}
\toprule
 & \textbf{CUDA} & \textbf{Triton} \\
\midrule
Top-5 models      & \gainhigh{+0.31} & \gainhigh{+0.29} \\
Bottom-5 models   & +0.02 & +0.05 \\
\bottomrule
\end{tabular}
\caption{Mean Speedup gains (\gainhigh{Green} denotes  $\geq$ 0.10) from \texttt{full} vs.\ \texttt{core} optimization.}
\label{tab:core_full_delta}
\end{table}

\subsection{Prompt Ablation}
\label{sec:prompt_ablation_results}

Table~\ref{tab:prompt_ablation} reports prompt ablations for GPT-5.5 at the \texttt{full} level, comparing the complete prompt with variants that remove either hardware context from the \textsc{Gpu} block or workload context from the \textsc{Data} block. The results show that while hardware-aware prompting provides clear benefits, workload-aware prompting is significantly more critical. In both frameworks, omitting data characteristics degrades performance more severely than removing GPU details. This impact is most pronounced under CUDA, where excluding data properties triggers a substantial performance drop of \textbf{$-$0.40} compared to \textbf{$-$0.27} for GPU details. Triton exhibits a parallel trend, showing only a minor degradation (\textbf{$-$0.06}) when omitting GPU specifications, but a sharper drop (\textbf{$-$0.22}) without workload data. Thus, concrete data properties such as table sizes and types are more vital for guiding effective kernel optimization than raw hardware specifications.

\begin{table}[t]
\centering
\small
\setlength{\tabcolsep}{8pt}
\begin{tabular}{@{}lrrr@{}}
\toprule
\textbf{Config} & \textbf{Base} & \textbf{No GPU Prompt} & \textbf{No Data Prompt} \\
\midrule
CUDA   & 2.11 & \drop{$-$0.27} & \drop{$-$0.40} \\
Triton & 2.05 & $-$0.06        & \drop{$-$0.22} \\
\bottomrule
\end{tabular}
\caption{Prompt ablation for GPT-5.5 (\texttt{full}) showing change in speedups. \drop{Red} denotes a drop $\geq$ 0.10.}
\label{tab:prompt_ablation}
\end{table}

\begin{figure}[t]
    \centering
    \includegraphics[width=0.9\columnwidth]{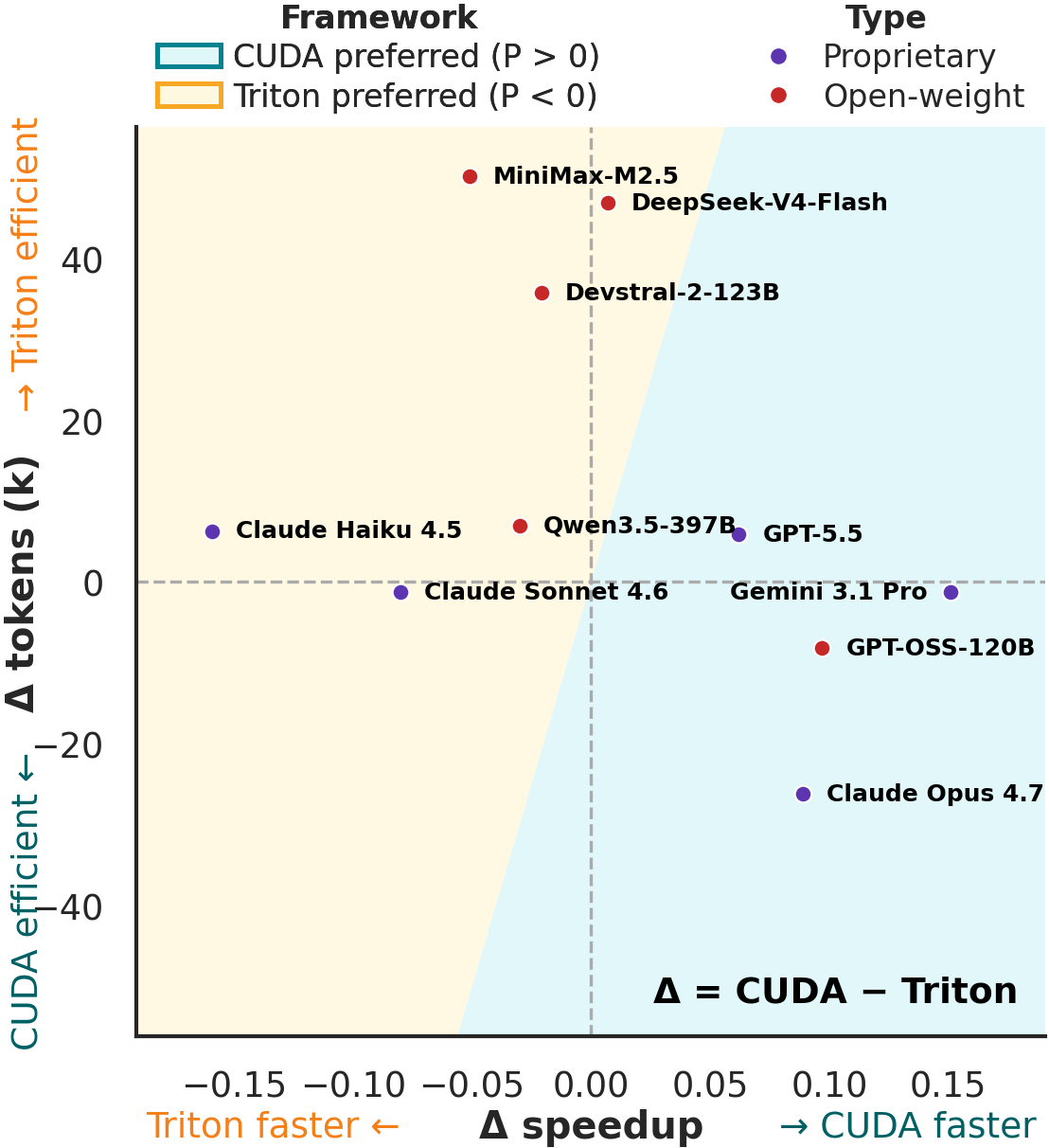}
    \caption{CUDA vs.\ Triton preference per model.
    $\Delta\text{speedup}$ and $\Delta\text{tokens}$ (k) measure the runtime
    and token usage change between frameworks. The boundary
    separates CUDA-preferred ($P>0$) from Triton-preferred ($P<0$) regions
    at default $\lambda=10$.}
    \label{fig:cuda_vs_triton}
\end{figure}

\subsection{CUDA vs.\ Triton Trade-offs}
\label{sec:cuda_triton_results}

Figure~\ref{fig:cuda_vs_triton} compares CUDA and Triton across models along two dimensions: runtime gain and generation cost. We define $\Delta$speedup as the speedup difference between CUDA and Triton, and $\Delta$tokens  as the corresponding difference in token usage measured in thousands. To combine them into a practical preference score, we use
\[
P = \lambda \,(100 \cdot \Delta\text{speedup}) - \Delta\text{tokens},
\]
where $\lambda$ denotes the maximum additional thousand tokens a practitioner is willing to spend for a 1\% speedup improvement. We use $\lambda=10$ as the default; models with $P>0$ favor CUDA, while models with $P<0$ favor Triton.

The resulting picture is mixed rather than absolute. Several proprietary models, including GPT-5.5, Gemini 3.1 Pro, and Claude Opus 4.7, fall in the CUDA-preferred region, indicating that CUDA offers better runtime at lower or comparable token cost. Several open-weight models, including MiniMax-M2.5, DeepSeek-V4-Flash, and Devstral-2-123B, fall in the Triton-preferred region. Qwen3.5-397B-A17B sits close to the decision boundary, making it the most backend-agnostic open-weight option in our study.

Overall, the results suggest a practical rule of thumb: CUDA provides the highest performance ceiling for the strongest models, while Triton is often a better cost-performance choice for weaker or open-weight models.

\section{Conclusion}

We introduced \textbf{DataKernelBench} for evaluating LLM-generated GPU kernels for analytical query processing. Across ten LLMs on TPC-H SF10, GPT-5.5 with CUDA-\texttt{full} achieves \textbf{2.11$\times$} over compiled TorchPlan with \textbf{100\%} pass rate. The gains are query-dependent: plan inspection identifies kernel fusion, fused aggregation, and broader execution-strategy changes, while generic GPU engines remain competitive on more complex queries. To process data beyond one GPU's memory, a Dask-cuDF proof of concept executes \textbf{TPC-H SF100} in on-demand partitions distributed across four H100 GPUs; all 22 queries pass, with \textbf{2.54$\times$} speedup over its partitioned TorchPlan baseline. These results establish LLM-generated kernels as a useful optimization path for selected recurring queries and DataKernelBench as a testbed for future work on LLM-driven database and GPU optimization.

\section{Limitations}

DataKernelBench is a first benchmark for this setting, and its current scope is intentionally controlled. 
First, the primary evaluation uses one fully instrumented setting: TPC-H at scale factor 10 on a single H100 GPU. Section~\ref{sec:scalability} provides preliminary evidence at SF100 on four H100 GPUs, but broader evaluation across workloads, hardware generations, and data distributions is needed before drawing general conclusions about analytical processing settings. 
Second, our framework is centered on the TorchPlan execution model implemented in PyTorch, with Triton and CUDA as the main synthesis targets. Other acceleration stacks and programming interfaces, such as Numba, CuTe DSL, or emerging domain-specific GPU frameworks, are not yet included. 
Third, the primary pipeline assumes that the working set fits within available device memory. The proof of concept relaxes this assumption through partitioned multi-GPU execution, but does not systematically evaluate spilling, unified memory, or scaling behavior across cluster sizes. 
Finally, baseline TorchPlans are generated and then validated before benchmarking. This design is appropriate for studying kernel synthesis conditioned on a fixed tensor program, but it means that DataKernelBench does not evaluate the full path from arbitrary SQL input to a final optimized kernel.

\section*{Acknowledgments}
This work is part of the ARMADA project (\url{https://armada-dn.eu/}) under the Marie Skłodowska-Curie Actions Doctoral Networks. The project is jointly funded by the Swiss State Secretariat for Education, Research and Innovation (SERI) under SBFI No.~24.00154 and by the European Union Horizon Europe research and innovation programme under Grant Agreement No.~101168951.

Views and opinions expressed are however those of the author(s) only and do not necessarily reflect those of the Swiss State Secretariat, the European Union, or the European Commission. Neither the Swiss State Secretariat, nor the European Union, nor the granting authorities can be held responsible for them.

\par\mbox{}\vfill\pagebreak
\bibliography{custom}

@inproceedings{kernelbench,
  title={{KernelBench}: Can {LLM}s Write Efficient {GPU} Kernels?},
  author={Ouyang, Anne and Guo, Simon and Arora, Simran and Zhang, Alex L and Hu, William and R{\'e}, Christopher and Mirhoseini, Azalia},
  booktitle={Proceedings of the 42nd International Conference on Machine Learning},
  series={Proceedings of Machine Learning Research},
  volume={267},
  pages={47356--47415},
  publisher={PMLR},
  year={2025},
  url={https://proceedings.mlr.press/v267/ouyang25a.html}
}

@inproceedings{tritonbench,
  title={Tritonbench: Benchmarking large language model capabilities for generating triton operators},
  author={Li, Jianling and Li, Shangzhan and Gao, Zhenye and Shi, Qi and Li, Yuxuan and Wang, Zefan and Huang, Jiacheng and WangHaojie, WangHaojie and Wang, Jianrong and Han, Xu and others},
  booktitle={Findings of the Association for Computational Linguistics: ACL 2025},
  pages={23053--23066},
  year={2025}
}

@misc{multikernelbench,
      title={MultiKernelBench: A Multi-Platform Benchmark for Kernel Generation}, 
      author={Zhongzhen Wen and Yinghui Zhang and Zhong Li and Zhongxin Liu and Linna Xie and Tian Zhang},
      year={2025},
      eprint={2507.17773},
      archivePrefix={arXiv},
      primaryClass={cs.DC},
      url={https://arxiv.org/abs/2507.17773}, 
}

@article{glm,
  title={Glm-4.5: Agentic, reasoning, and coding (arc) foundation models},
  author={Zeng, Aohan and Lv, Xin and Zheng, Qinkai and Hou, Zhenyu and Chen, Bin and Xie, Chengxing and Wang, Cunxiang and Yin, Da and Zeng, Hao and Zhang, Jiajie and others},
  journal={arXiv preprint arXiv:2508.06471},
  year={2025}
}

@misc{gptoss,
      title={gpt-oss-120b \& gpt-oss-20b Model Card}, 
      author={OpenAI},
      year={2025},
      eprint={2508.10925},
      archivePrefix={arXiv},
      primaryClass={cs.CL},
      url={https://arxiv.org/abs/2508.10925}, 
}

@techreport{qwen3_coder_next,
  title        = {Qwen3-Coder-Next Technical Report},
  author       = {{Qwen Team}},
  url          = {https://github.com/QwenLM/Qwen3-Coder/blob/main/qwen3_coder_next_tech_report.pdf},
  note         = {Accessed: 2026-03-13},
  year         = {2026}
}

@misc{qwen3.5,
    title  = {{Qwen3.5}: Towards Native Multimodal Agents},
    author = {{Qwen Team}},
    month  = {February},
    year   = {2026},
    url    = {https://qwen.ai/blog?id=qwen3.5}
}

@misc{deepseekv4,
      title={DeepSeek-V4: Towards Highly Efficient Million-Token Context Intelligence},
      author={DeepSeek-AI},
      year={2026},
}

@misc{MiniMax-M2.5,
    author = {MiniMax AI},
    title = {MiniMax-M2.5: A Professional Employee and SOTA Model for Agentic Workflows},
    year = {2026},
    publisher = {Hugging Face},
    journal = {Hugging Face Repository},
    howpublished = {\url{https://huggingface.co/MiniMaxAI/MiniMax-M2.5}} 
}

@article{devstral,
  title={Devstral: Fine-tuning Language Models for Coding Agent Applications},
  author={Rastogi, Abhinav and Yang, Adam and Jiang, Albert Q and Liu, Alexander H and Sablayrolles, Alexandre and H{\'e}liou, Am{\'e}lie and Martin, Am{\'e}lie and Agarwal, Anmol and Ehrenberg, Andy and Lo, Andy and others},
  journal={arXiv preprint arXiv:2509.25193},
  year={2025}
}

@misc{tpch,
  title = {TPC Benchmark H (TPC-H)},
  author = {{Transaction Processing Council}},
  year = {2018},
  howpublished = {\url{https://www.tpc.org/tpch/}}
}

@software{pandas,
  author = {{Pandas}},
  title = {pandas-dev/pandas: Pandas},
  year = {2020},
  publisher = {Zenodo},
  doi = {10.5281/zenodo.3509134},
  url = {https://doi.org/10.5281/zenodo.3509134}
}

@inproceedings{dask,
  title={Dask: Parallel computation with blocked algorithms and task scheduling},
  author={Rocklin, Matthew},
  booktitle={Proceedings of the 14th Python in Science Conference},
  pages={130--136},
  year={2015}
}

@misc{cudf,
  title = {RAPIDS cuDF: GPU DataFrame Library},
  author = {{RAPIDS}},
  year = {2023},
  howpublished = {\url{https://rapids.ai}}
}

@misc{velox,
  title={Velox: A unified execution engine for data management systems},
  author={{Velox}},
  year={2022},
  howpublished={\url{https://github.com/facebookincubator/velox}}
}

@inproceedings{duckdb,
  title={DuckDB: an embeddable analytical database},
  author={Raasveldt, Mark and M{\"u}hleisen, Hannes},
  booktitle={Proceedings of the 2019 International Conference on Management of Data},
  pages={1981--1984},
  year={2019}
}

@article{bespoke_olap,
  title={Bespoke OLAP: Synthesizing Workload-Specific One-size-fits-one Database Engines},
  author={Wehrstein, Johannes and Eckmann, Timo and Jasny, Matthias and Binnig, Carsten},
  journal={arXiv preprint arXiv:2603.02001},
  year={2026}
}

@article{tqp1,
  title={Query Processing on Tensor Computation Runtimes},
  author={He, Dong and Nakandala, Supun and Banda, Dalitso and Sen, Rathijit and Saur, Karla and Park, Kwanghyun and Curino, Carlo and Camacho-Rodr{\'\i}guez, Jes{\'u}s and Karanasos, Konstantinos and Interlandi, Matteo},
  journal={Proceedings of the VLDB Endowment},
  volume={15},
  number={11},
  pages={2811--2825},
  year={2022},
  doi={10.14778/3551793.3551833},
  url={https://doi.org/10.14778/3551793.3551833}
}

@article{tqp2,
  title={Share the Tensor Tea: How Databases Can Leverage the Machine Learning Ecosystem},
  author={Asada, Yuki and Fu, Victor and Gandhi, Apurva and Gemawat, Advitya and Zhang, Lihao and He, Dong and Gupta, Vivek and Nosakhare, Ehi and Banda, Dalitso and Sen, Rathijit and Interlandi, Matteo},
  journal={Proceedings of the VLDB Endowment},
  volume={15},
  number={12},
  pages={3598--3601},
  year={2022},
  doi={10.14778/3554821.3554853},
  url={https://doi.org/10.14778/3554821.3554853}
}

@article{tqp_multi_gpu,
  title={Terabyte-Scale Analytics in the Blink of an Eye},
  author={Wu, Bowen and Cui, Wei and Curino, Carlo and Interlandi, Matteo and Sen, Rathijit},
  journal={Proceedings of the VLDB Endowment},
  volume={19},
  number={2},
  pages={141--155},
  year={2025},
  doi={10.14778/3773749.3773754},
  url={https://doi.org/10.14778/3773749.3773754}
}

@article{tritonrl,
  title={Tritonrl: Training llms to think and code triton without cheating},
  author={Woo, Jiin and Zhu, Shaowei and Nie, Allen and Jia, Zhen and Wang, Yida and Park, Youngsuk},
  journal={arXiv preprint arXiv:2510.17891},
  year={2025}
}

@article{cuda_kernel_agent,
  title={CUDA Agent: Large-Scale Agentic RL for High-Performance CUDA Kernel Generation},
  author={Dai, Weinan and Wu, Hanlin and Yu, Qiying and Gao, Huan-ang and Li, Jiahao and Jiang, Chengquan and Lou, Weiqiang and Song, Yufan and Yu, Hongli and Chen, Jiaze and others},
  journal={arXiv preprint arXiv:2602.24286},
  year={2026}
}

@article{kernelevolve,
  title={Kernelevolve: Scaling agentic kernel coding for heterogeneous ai accelerators at meta},
  author={Liao, Gang and Qin, Hongsen and Wang, Ying and Golden, Alicia and Kuchnik, Michael and Yetim, Yavuz and Ang, Jia Jiunn and Fu, Chunli and He, Yihan and Hsia, Samuel and others},
  journal={arXiv preprint arXiv:2512.23236},
  year={2025}
}

@article{lange2025towards,
  title={Towards robust agentic {CUDA} kernel benchmarking, verification, and optimization},
  author={Lange, Robert Tjarko and Sun, Qi and Prasad, Aaditya and Faldor, Maxence and Tang, Yujin and Ha, David},
  journal={arXiv preprint arXiv:2509.14279},
  year={2025}
}

@inproceedings{baronio2026kevin,
  title={Kevin: Multi-turn {RL} for generating {CUDA} kernels},
  author={Baronio, Carlo and Marsella, Pietro and Pan, Ben and Guo, Simon and Alberti, Silas},
  booktitle={International Conference on Learning Representations},
  volume={2026},
  pages={83418--83452},
  year={2026}
}

@article{openai_gpt4,
  title={GPT-4 Technical Report},
  author={{OpenAI}},
  journal={arXiv preprint arXiv:2303.08774},
  year={2023},
  url={https://arxiv.org/abs/2303.08774}
}

@misc{gpt55,
  author    = {OpenAI},
  title     = {Introducing {GPT}-5.5},
  year      = {2026},
  month     = {April},
  howpublished = {\url{https://openai.com/index/introducing-gpt-5-5/}},
  note      = {Accessed: 2026-05-26}
}

@article{anthropic_claude,
  title={Constitutional AI: Harmlessness from AI Feedback},
  author={Bai, Yuntao and Kadavath, Saurabh and Kundu, Saurav and Askell, Amanda and others},
  journal={arXiv preprint arXiv:2212.08073},
  year={2022},
  url={https://arxiv.org/abs/2212.08073}
}

@misc{anthropic2025haiku45,
  author       = {Anthropic},
  title        = {Introducing {Claude} {Haiku} 4.5},
  year         = {2025},
  month        = {October},
  howpublished = {\url{https://www.anthropic.com/news/claude-haiku-4-5}},
  note         = {Accessed: 2026-05-26}
}

@misc{anthropic2026sonnet46,
  author       = {Anthropic},
  title        = {Introducing {Claude} {Sonnet} 4.6},
  year         = {2026},
  month        = {February},
  howpublished = {\url{https://www.anthropic.com/news/claude-sonnet-4-6}},
  note         = {Accessed: 2026-05-26}
}

@misc{anthropic2026opus47,
  author       = {Anthropic},
  title        = {Introducing {Claude} {Opus} 4.7},
  year         = {2026},
  month        = {April},
  howpublished = {\url{https://www.anthropic.com/news/claude-opus-4-7}},
  note         = {Accessed: 2026-05-26}
}

@misc{google2026gemini31pro,
  author       = {Google DeepMind},
  title        = {{Gemini} 3.1 {Pro}: A smarter model for your most complex tasks},
  year         = {2026},
  month        = {February},
  howpublished = {\url{https://blog.google/innovation-and-ai/models-and-research/gemini-models/gemini-3-1-pro/}},
  note         = {Accessed: 2026-05-26}
}

@article{lqs,
  title={Learned Query Superoptimization},
  author={Marcus, Ryan},
  journal={arXiv preprint arXiv:2303.15308},
  year={2023}
}

@misc{idc_ai_infrastructure_2025,
  title = {{IDC}: Artificial Intelligence Infrastructure Spending to Reach \$758Bn by 2029},
  author = {{International Data Corporation (IDC)}},
  year = {2025},
  url = {https://www.channel-impact.com/idc-artificial-intelligence-infrastructure-spending-to-reach-758bn-by-2029/}
}

@misc{gartner_ai_spending_2025,
    title = {Gartner Says Worldwide {AI} Spending Will Total \$1.5 Trillion in 2025},
    author = {{Gartner, Inc.}},
    year = {2025},
    url = {https://www.gartner.com/en/newsroom/press-releases/2025-09-17-gartner-says-worldwide-ai-spending-will-total-1-point-5-trillion-in-2025}
}

@inproceedings{learned_index,
  title={The case for learned index structures},
  author={Kraska, Tim and Beutel, Alex and Chi, Ed H and Dean, Jeffrey and Polyzotis, Neoklis},
  booktitle={Proceedings of the 2018 International Conference on Management of Data (SIGMOD)},
  pages={489--504},
  year={2018}
}

@article{neo_optimizer,
  title={Neo: A learned query optimizer},
  author={Marcus, Ryan and Negi, Parimarjan and Mao, Hongzi and Zhang, Chi and Alizadeh, Mohammad and Kraska, Tim and Papaemmanouil, Olga and Tatbul, Nesime},
  journal={Proceedings of the VLDB Endowment},
  volume={12},
  number={11},
  pages={1705--1718},
  year={2019}
}

@article{bird_text2sql,
  title={Can LLM already serve as a database interface? A BIg Data Benchmark for Text-to-SQL},
  author={Li, Jinyang and Hui, Binyuan and Qu, Ge and Li, Binyuan and Yang, Jiaxi and Li, Bowen and Wang, Bailin and Qin, Bowen and Geng, Ruiying and Huo, Nan and others},
  journal={Advances in Neural Information Processing Systems (NeurIPS)},
  volume={36},
  year={2024}
}

@article{dail_sql,
  title={Text-to-SQL Empowered by Large Language Models: A Benchmark Evaluation},
  author={Gao, Dawei and Wang, Haibin and Li, Yaliang and Sun, Xiuyu and Qian, Yichen and Ding, Bolin and Zhou, Jingren},
  journal={Proceedings of the VLDB Endowment},
  volume={17},
  number={5},
  pages={1132--1145},
  year={2024}
}

@misc{kinetica,
  author = {{Kinetica}},
  title = {{Kinetica}},
  note = {Accessed: 2026-03-13},
  year = {2026},
  url = {https://www.kinetica.com/}
}

@misc{sqream,
  author = {{SQream Technologies}},
  title = {{SQream DB}},
  note = {Accessed: 2026-03-13},
  year = {2026},
  url = {https://sqream.com/product/sqreamdb/}
}

@article{evalplusperformance,

title={Evaluating language models for efficient code generation},

author={Liu, Jiawei and Xie, Songrun and Wang, Junhao and Wei, Yuxiang and Ding, Yifeng and Zhang, Lingming},

journal={arXiv preprint arXiv:2408.06450},

year={2024}

}

@article{livecodebench,

title={Livecodebench: Holistic and contamination free evaluation of large language models for code},

author={Jain, Naman and Han, King and Gu, Alex and Li, Wen-Ding and Yan, Fanjia and Zhang, Tianjun and Wang, Sida and Solar-Lezama, Armando and Sen, Koushik and Stoica, Ion},

journal={arXiv preprint arXiv:2403.07974},

year={2024}

}

@inproceedings{sirius,
  title={Rethinking Analytical Processing in the GPU Era},
  author={Yogatama, Bobbi and Yang, Yifei and Kristensen, Kevin and Sarda, Devesh and Kim, Abigale and Cockcroft, Adrian and Teng, Yu and Patterson, Joshua and Kimball, Gregory and McKinney, Wes and Gong, Weiwei and Yu, Xiangyao},
  booktitle={Conference on Innovative Data Systems Research (CIDR)},
  year={2026},
  url={https://www.vldb.org/cidrdb/2026/rethinking-analytical-processing-in-the-gpu-era.html}
}

@article{pystachio,
  title={PystachIO: Efficient Distributed GPU Query Processing with PyTorch over Fast Networks \& Fast Storage},
  author={Luo, Jigao and Boeschen, Nils and El-Hindi, Muhammad and Binnig, Carsten},
  journal={arXiv preprint arXiv:2512.02862},
  year={2025}
}

@inproceedings{kernel_weaver,
  title={Kernel weaver: Automatically fusing database primitives for efficient gpu computation},
  author={Wu, Haicheng and Diamos, Gregory and Cadambi, Srihari and Yalamanchili, Sudhakar},
  booktitle={2012 45th Annual IEEE/ACM International Symposium on Microarchitecture},
  pages={107--118},
  year={2012},
  organization={IEEE}
}

@inproceedings{tpch_analysis,
author = {Boncz, Peter and Neumann, Thomas and Erling, Orri},
title = {TPC-H Analyzed: Hidden Messages and Lessons Learned from an Influential Benchmark},
year = {2013},
isbn = {9783319049359},
publisher = {Springer-Verlag},
address = {Berlin, Heidelberg},
url = {https://doi.org/10.1007/978-3-319-04936-6_5},
doi = {10.1007/978-3-319-04936-6_5},
booktitle = {Revised Selected Papers of the 5th TPC Technology Conference on Performance Characterization and Benchmarking - Volume 8391},
pages = {61–76},
numpages = {16}
}

@article{tqex,
author = {Zhang, Haitao and Pang, Ran and Zhu, Yuanyuan and Zhang, Hao and Gao, Congli and Zhong, Ming and Jiang, Jiawei and Qian, Tieyun and Yu, Jeffrey Xu},
title = {TQEx: Tensor-based Query Engine Enhanced by Bridging the Gap},
year = {2025},
issue_date = {December 2025},
publisher = {Association for Computing Machinery},
address = {New York, NY, USA},
volume = {3},
number = {6},
url = {https://doi.org/10.1145/3769835},
doi = {10.1145/3769835},
journal = {Proc. ACM Manag. Data},
month = dec,
articleno = {370},
numpages = {27}
}

@article{tqp_compressed,
  title={GPU Acceleration of SQL Analytics on Compressed Data},
  author={Huang, Zezhou and Sakowski, Krystian and Lehnert, Hans and Cui, Wei and Curino, Carlo and Interlandi, Matteo and Dumitru, Marius and Sen, Rathijit},
  journal={Proceedings of the VLDB Endowment},
  volume={19},
  number={3},
  pages={320--333},
  year={2025},
  publisher={VLDB Endowment},
  doi={10.14778/3778092.3778095},
  url={https://doi.org/10.14778/3778092.3778095}
}

@article{team2024falcon,
  title={The falcon 3 family of open models, December 2024},
  author={Team, Falcon-LLM},
  journal={URL https://huggingface. co/blog/falcon3},
  year={2024}
}

\clearpage
\appendix

\begin{figure}[!h]
    \centering
    \includegraphics[width=1\columnwidth]{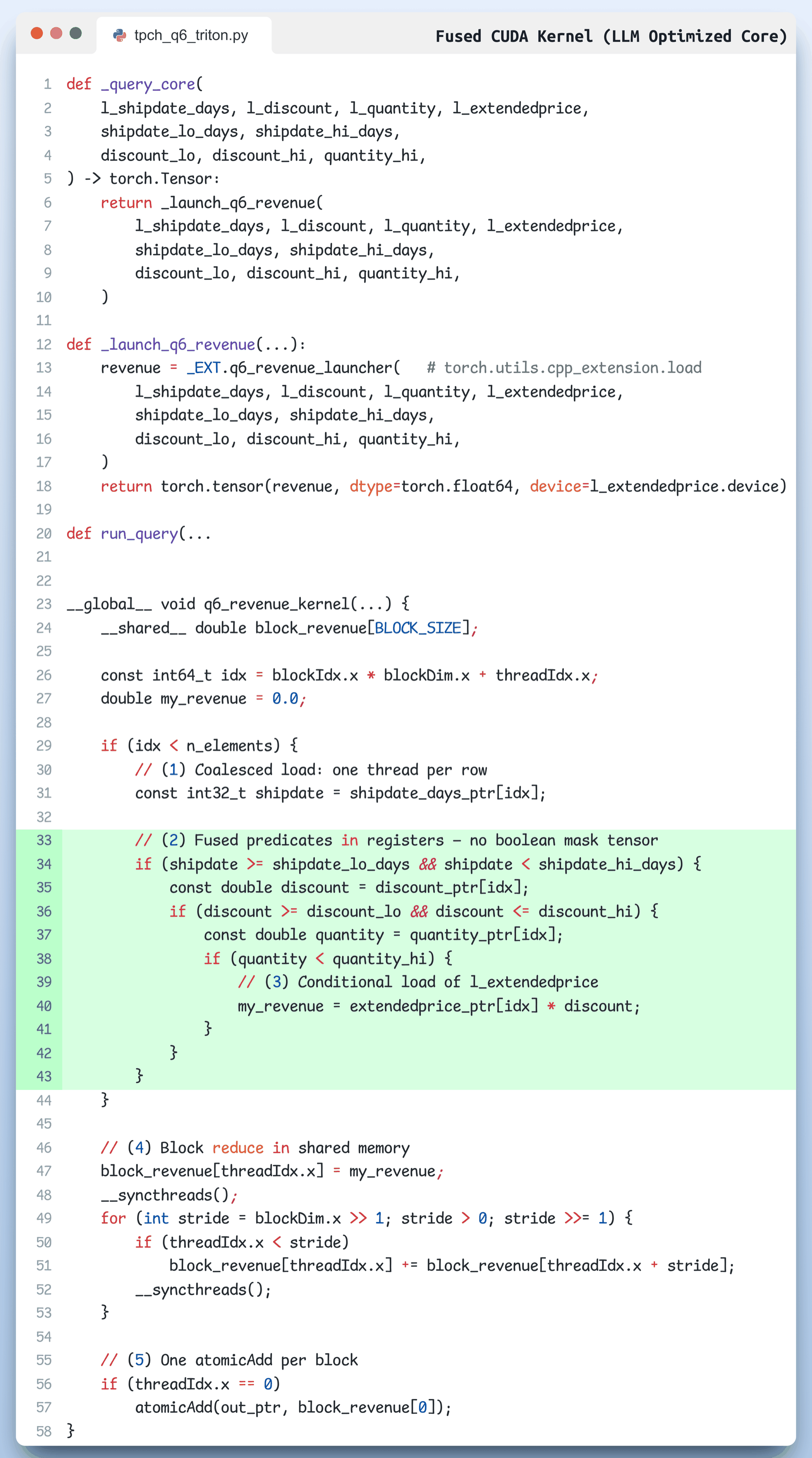}
    \caption{TPC-H Q6 Example: Fused CUDA kernel.
      Like the Triton variant in Figure~\ref{fig:translation}(C), \texttt{\_query\_core} delegates to a compiled launcher whose \texttt{\_\_global\_\_} kernel fuses filter, projection, and aggregation in one pass over \texttt{lineitem}.
      Nested branch predicates avoid materializing a boolean mask; \texttt{l\_extendedprice} is loaded only for rows that pass all filters; partial sums are reduced in shared memory with one \texttt{atomicAdd} per block.
    }
    \label{app:fig:translation_cuda}
\end{figure}

\begin{figure}[!h]
    \centering
    \includegraphics[width=1\columnwidth]{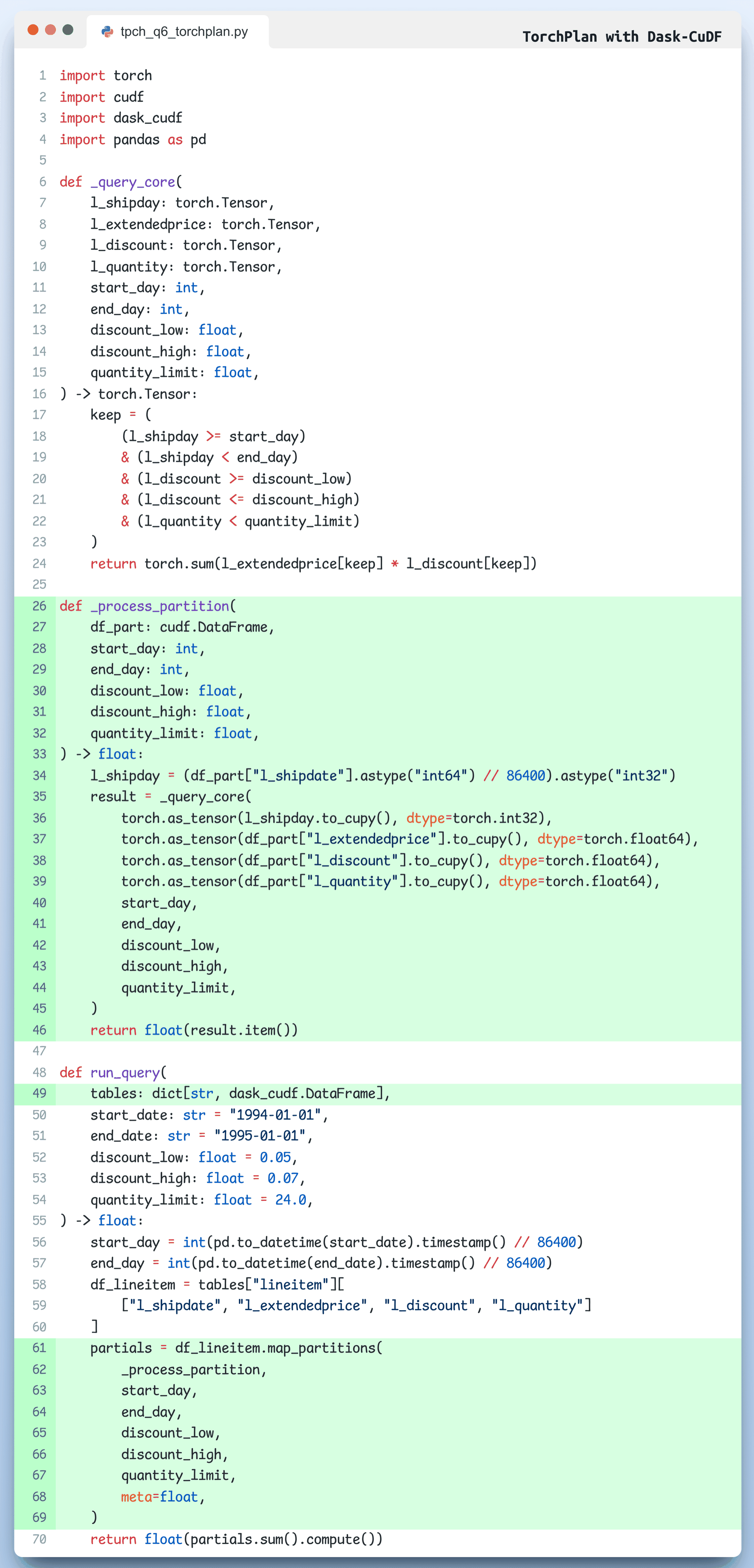}
    \caption{TPC-H Q6 Example: TorchPlan with Dask-cuDF.
      Like the eager-cuDF TorchPlan in Figure~\ref{fig:translation}(B), \texttt{\_query\_core} implements the tensor hot path. The highlighted \texttt{\_process\_partition} converts each cuDF chunk to tensors and calls \texttt{\_query\_core}; \texttt{run\_query} uses Dask-cuDF \texttt{map\_partitions} to run those chunks on available GPU workers and sums the partial results.
    }
    \label{app:fig:translation_dask}
\end{figure}

\begin{figure}[!h]
    \centering
    \includegraphics[width=0.95\columnwidth]{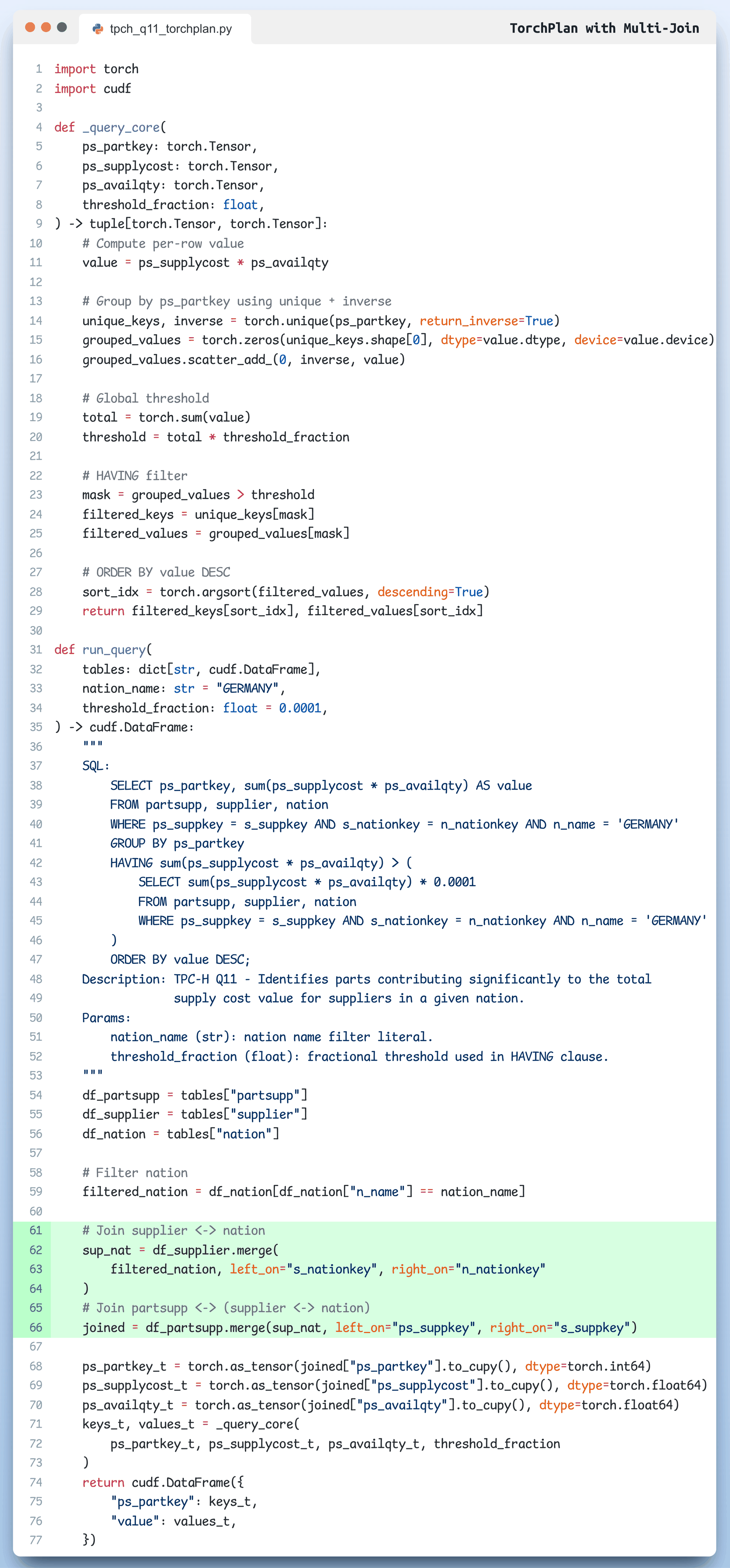}
    \caption{TPC-H Q11 Example: Multi-join TorchPlan.
      Complementing the single-table Q6 example, \texttt{run\_query} handles cuDF table access, nation filtering, and the highlighted multi-table joins before converting joined columns to tensors; \texttt{\_query\_core} aggregates supply cost by part key, applies the HAVING threshold, and sorts.
      The \texttt{core} task preserves this boundary, whereas \texttt{full} may change the join strategy and how work is divided between the two functions.
    }
    \label{app:fig:multi_join_torchplan}
\end{figure}

\section{Fused CUDA Kernel Example}
\label{app:cuda_example}
Figure~\ref{app:fig:translation_cuda} shows the GPT-5.5 CUDA variant for the same TPC-H Q6 query as Figure~\ref{fig:translation}. It uses the same TorchPlan interface: \texttt{run\_query} handles cuDF I/O and parameters, while \texttt{\_query\_core} delegates to a compiled launcher and fused device kernel. We include this variant in the appendix because the main paper highlights the Triton core snippet in Figure~\ref{fig:translation}(C).

\section{Partitioned Dask-cuDF TorchPlan Example}
\label{app:dask_example}
Figure~\ref{app:fig:translation_dask} shows the partitioned TorchPlan for the same TPC-H Q6 query as Figure~\ref{fig:translation}(B). The tensor hot path is unchanged; only table handling moves from eager cuDF to Dask-cuDF so that partitions can execute across GPUs. Section~\ref{sec:scalability} uses this interface for SF100.

\section{Multi-Join TorchPlan Example}
\label{app:multi_join_example}

Figure~\ref{app:fig:multi_join_torchplan} shows the validated TorchPlan for TPC-H Q11. As in the Q6 CUDA example in Figure~\ref{app:fig:translation_cuda}, the interface separates table handling from the tensor hot path: joins remain in \texttt{run\_query}, while grouping, thresholding, and sorting run in \texttt{\_query\_core}.

\section{TorchPlan Generator Sensitivity}
\label{app:torchplan_sensitivity}

Claude Opus 4.7 was used for the main benchmark because it produced a complete validated set of \textbf{22/22 TorchPlans}; MiniMax M2.5 produced 20/22 under the same protocol. For the sensitivity analysis, GPT-5.5 also generated \textbf{22/22 valid TorchPlans}, but their compiled workload runtime was 1.74\,s, approximately 15\% slower than the 1.51\,s runtime of the Opus 4.7 TorchPlans. Table~\ref{tab:torchplan_sensitivity} compares baseline and optimized runtimes across both sets.

\begin{table}[t]
\centering
\footnotesize
\setlength{\tabcolsep}{3pt}
\begin{tabular}{@{}lrrr@{}}
\toprule
\textbf{Method} & \multicolumn{2}{c}{\textbf{TorchPlan generator}} & \textbf{Change} \\
\cmidrule(lr){2-3}
& \textbf{Opus 4.7} & \textbf{GPT-5.5} & \\
\midrule
\multicolumn{4}{@{}l}{\textit{Baseline}} \\
TorchPlan-\texttt{compile} & 1.51\,s & 1.74\,s & +15\% \\
\addlinespace
\multicolumn{4}{@{}l}{\textit{Optimized implementations}} \\
GPT-5.5  & 0.71\,s & 0.89\,s & +25\% \\
Opus 4.7 & 1.00\,s & 0.91\,s & -9\% \\
\bottomrule
\end{tabular}
\caption{Total workload runtime (s) for TorchPlans from two generators. Optimized rows use CUDA-\texttt{full}; changes are approximate and relative to the Opus 4.7 TorchPlans. Lower is better.}
\label{tab:torchplan_sensitivity}
\end{table}

Absolute runtimes change, but both optimized implementations remain faster than their corresponding baselines, and GPT-5.5 remains the strongest optimizer. Each optimizer is fastest on TorchPlans generated by the other strong model, suggesting that cross-model diversity may be useful. Overall, the main finding is not specific to TorchPlans generated by one model.

\section{Hardware and Software Environment}
\label{app:hardware-software-environment}

Table~\ref{tab:software-environment} reports the main software packages used for benchmark execution and kernel evaluation. Tables~\ref{tab:gpu-environment} and~\ref{tab:cpu-environment} report the GPU and CPU environments. The LLM-generated CUDA/Triton kernels and Sirius baseline use GPU acceleration, while DuckDB is used as a CPU-only database baseline.

\begin{table}[t]
\centering
\small
\begin{tabular}{ll}
\toprule
Package & Version \\
\midrule
Python & 3.13.11 \\
pandas & 2.3.3 \\
cuDF & 26.04.000 \\
PyTorch & 2.12.0+cu130 \\
Triton & 3.7.0 \\
CUDA & 13.0 \\
DuckDB & 1.5.2 \\
\bottomrule
\end{tabular}
\caption{Software environment used for DataKernelBench experiments.}
\label{tab:software-environment}
\end{table}

\begin{table}[t]
\centering
\small
\begin{tabular}{ll}
\toprule
Field & Value \\
\midrule
GPU & NVIDIA H100 SXM5 \\
Architecture & Hopper (GH100) \\
SMs & 132 \\
CUDA cores & 16,896 \\
Tensor Cores & 528, 4th gen \\
GPU memory & 80 GB HBM3 \\
Memory bandwidth & 3.35 TB/s \\
L2 cache & 50 MB \\
Peak FP64 & 34 TFLOPS \\
Peak FP32 & 67 TFLOPS \\
FP8 Tensor Core & Supported \\
TDP & Up to 700 W \\
Interconnect & NVLink 900 GB/s; \\
& PCIe Gen5 128 GB/s \\
Warp size & 32 \\
\bottomrule
\end{tabular}
\caption{GPU environment used for LLM-generated kernel evaluation and GPU-accelerated baselines.}
\label{tab:gpu-environment}
\end{table}

\begin{table}[!h]
\centering
\small
\begin{tabular}{ll}
\toprule
Field & Value \\
\midrule
CPU & Intel Xeon Platinum 8468 \\
Architecture & Sapphire Rapids; \\
& 4th Gen Xeon Scalable \\
Sockets & 2 \\
Cores & 96 total, 48 per socket \\
Logical CPUs & 96, SMT off \\
NUMA nodes & 2 \\
Base frequency & 2.10 GHz \\
Max turbo & 3.80 GHz \\
L3 cache & 210 MiB total \\
System memory & $\sim$2.0 TB DDR5 \\
TDP & 350 W per socket \\
Socket & FCLGA4677, 2S platform \\
CPU database baseline & DuckDB 1.5.2, CPU-only \\
\bottomrule
\end{tabular}
\caption{CPU environment used for CPU-side execution and the DuckDB CPU-only baseline.}
\label{tab:cpu-environment}
\end{table}

\section{Repair Behavior}
\label{app:repair_behavior}

We analyze \textbf{880 generation trajectories}: 22 queries $\times$ 10 models $\times$ 2 frameworks $\times$ 2 optimization levels on H100 at SF10.
\begin{itemize}
    \item \textbf{Repair effectiveness.} Overall, 77.5\% eventually produce a correct kernel with at least 1.05$\times$ speedup. Cumulative success rises from 40\% after round~1 to 57\% after round~2 and 65\% after round~3. Of the remaining trajectories, 17.5\% never become correct, while 5\% become correct but remain below the speedup threshold.
    \item \textbf{Failure categories.} Among round-1 failures, approximately 38\% fail to compile or import, 52\% crash or time out during execution, and 9\% return incorrect output, primarily wrong values or row counts; column mismatches account for less than 1\%.
    \item \textbf{Backend differences.} Approximately 74\% of CUDA failures are compilation or import errors, whereas 79\% of Triton failures are execution errors.
    \item \textbf{Query difficulty.} Q13 is the hardest query, with 13/40 successful trajectories. Q20, Q3, and Q1 are also difficult, while Q6 and Q19 succeed in all 40 trajectories.
    \item \textbf{Prompt improvements.} Recurring failures led us to pin runtime versions and add guidance on separate kernel source strings and epoch-day date representation, reducing repair rounds without materially changing final speedup.
\end{itemize}

\begin{figure}[t]
    \centering
    \includegraphics[width=1\columnwidth]{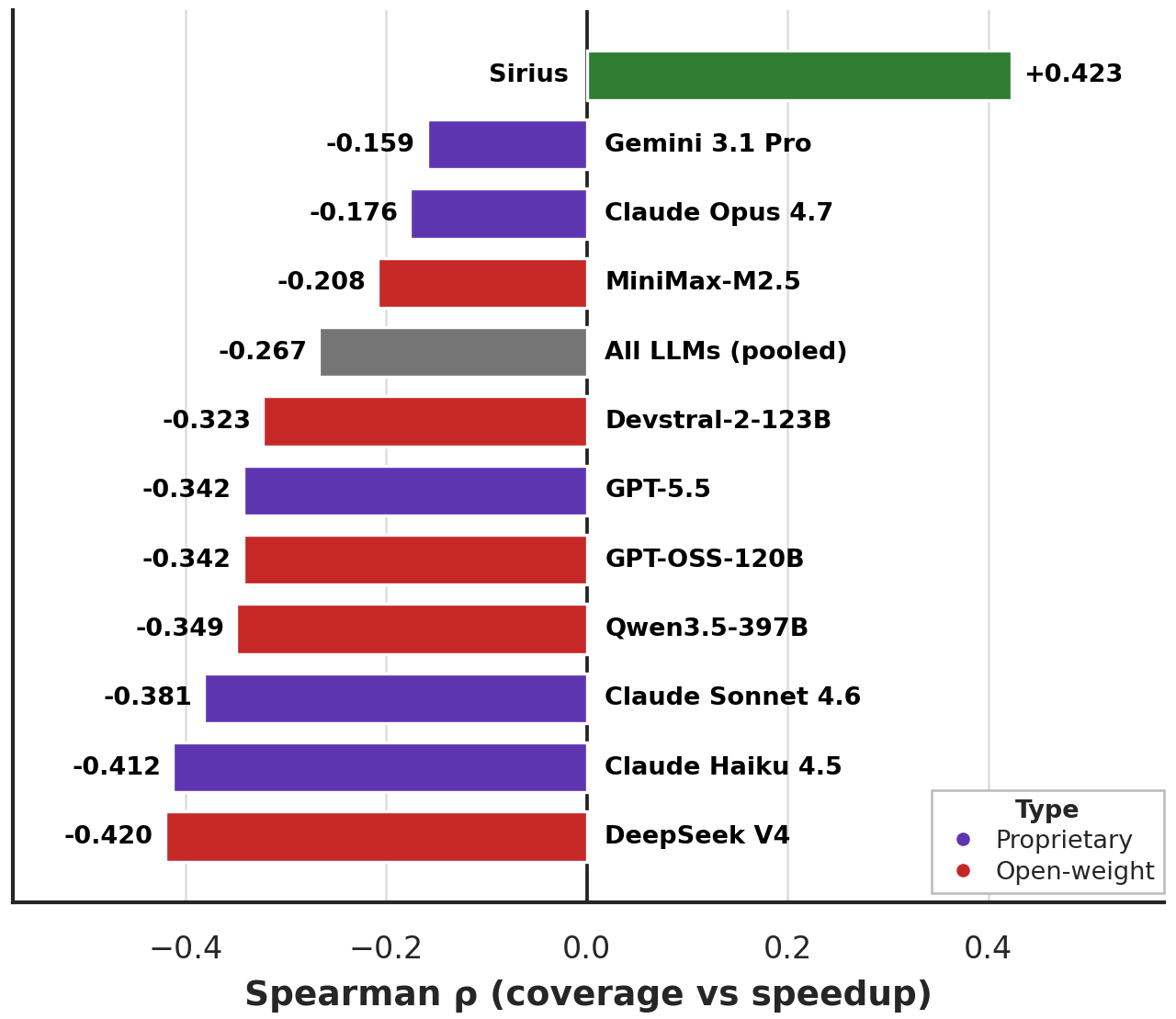}
    \caption{Spearman correlation between choke-point coverage and per-query speedup. Coverage counts the number of distinct choke-point categories assigned to each TPC-H query. All ten LLMs show negative correlation, while Sirius shows positive correlation}
    \label{fig:chokepoint-correlation}
\end{figure}

\section{Choke-Point Coverage Analysis}
\label{app:chokepoint-analysis}

To better understand when LLM-generated kernels outperform a general-purpose GPU database baseline, we define choke-point coverage as the number of distinct choke-point categories \citep{tpch_analysis} explicitly associated with each TPC-H query. We count only query-category assignments named in the taxonomy, making the measure conservative: queries may involve additional relational patterns that are not counted in our annotation.

For each system, averaged across 4 framework-optimization level combinations, we compute the Spearman rank correlation between the 22 per-query coverage values and the corresponding 22 per-query speedups. Figure~\ref{fig:chokepoint-correlation} shows that all ten LLMs have negative correlation between coverage and speedup, while Sirius has positive correlation. When pooling all LLM-query pairs, the trend remains negative with $\rho=-0.267$ ($n=220$). We also find that the head-to-head LLM/Sirius speedup ratio decreases with coverage for all ten LLMs. These results support the complementary interpretation in Section~\ref{sec:query_level_results}: current LLM-based specialization is most effective for narrower query patterns, while Sirius becomes relatively stronger on queries that combine more relational choke points.

\begin{figure}[t]
    \centering
    \includegraphics[width=1\columnwidth]{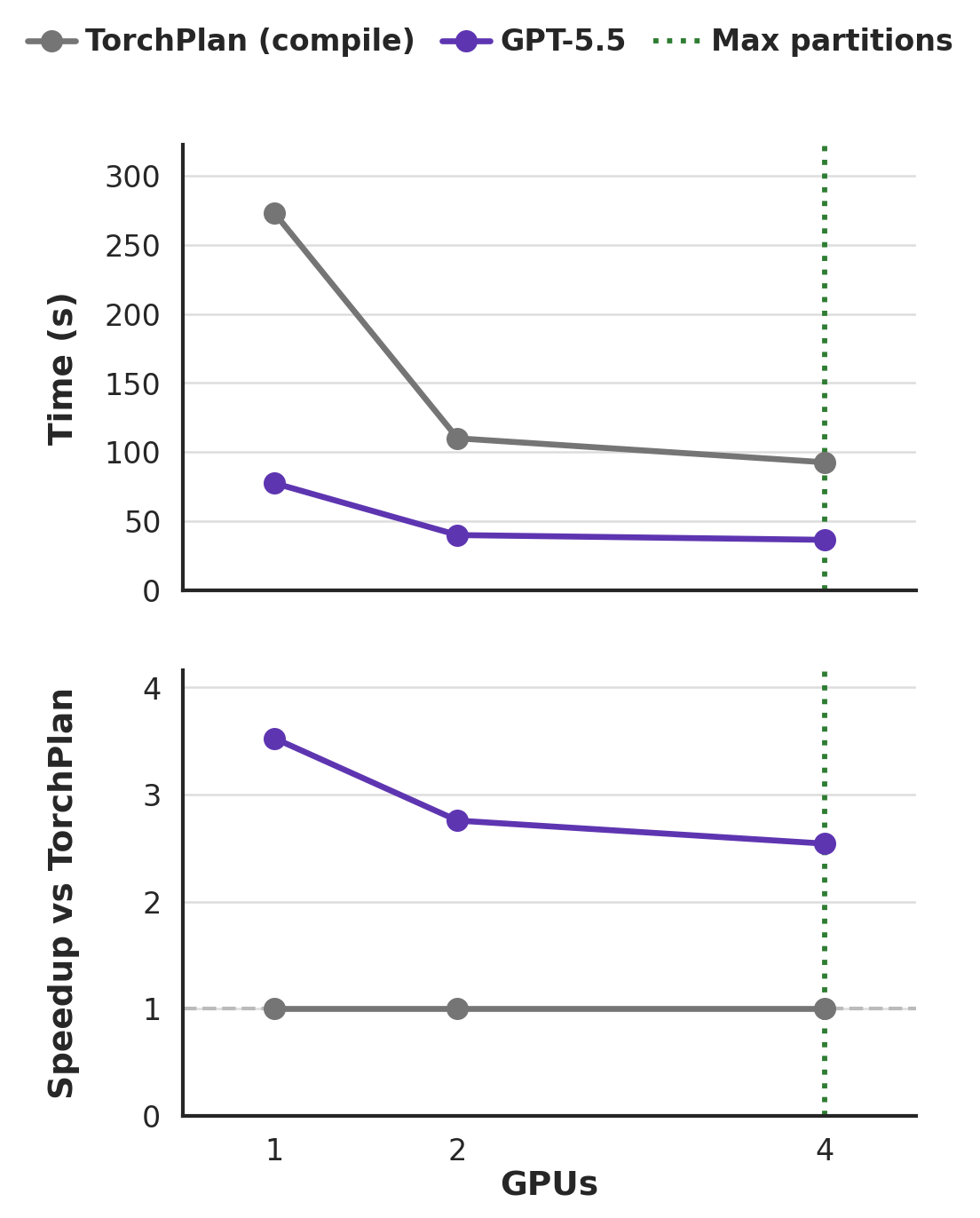}
    \caption{Multi-GPU scaling of GPT-5.5 CUDA-\texttt{full} kernel plans versus TorchPlan-\texttt{compile} on TPC-H SF100 (H100, Dask-cuDF, 10\,GB chunks). The green dotted line marks the widest table partition count (\texttt{lineitem} = 4). \textbf{Top:} runtime over all 22 queries. \textbf{Bottom:} overall speedup versus TorchPlan. The CUDA kernel plan advantage over TorchPlan is largest on one GPU ($3.53\times$) and is $2.54\times$ at 4 GPUs.}
    \label{app:fig:gpu_scaling}
\end{figure}

\section{Multi-GPU Scaling}
\label{app:gpu_scaling}

Figure~\ref{app:fig:gpu_scaling} varies the GPU count from 1 to 4 for the SF100 setting in Section~\ref{sec:scalability}. Dask-cuDF represents each table as partitions and, through \texttt{map\_partitions}, runs independent partitions concurrently on the available GPU workers, queuing any remaining partitions until a worker is free. Under this layout \texttt{lineitem} has four partitions and every other table has one, so four is the maximum number of partitions that can run at once. We therefore stop at four GPUs. As in Section~\ref{sec:exp_setup}, queries that fail or do not reach $1.05\times$ are credited at the TorchPlan-\texttt{compile} runtime. Runtime over all 22 queries falls from 1 to 4 GPUs for both the CUDA kernel plans (77.5\,s to 36.44\,s) and TorchPlan-\texttt{compile} (273\,s to 92.6\,s).

At higher scale factors, more partitions would be available, so additional GPUs could run more partitions in parallel. A possible further improvement is to rewrite Dask-cuDF multi-GPU orchestration with LLM-generated native PyTorch code.


\end{document}